\documentstyle{article}
\begin{document}
\bibliographystyle{plain}
\hyphenation{mono-tonicity Mono-tonicity mono-tonic Mono-tonic Mo-notonicity
mo-notonicity supra-classical}
\newtheorem{theorem}{Theorem}[section]
\newtheorem{corollary}{Corollary}[section]
\newtheorem{lemma}{Lemma}[section]
\newtheorem{exercise}{Exercise}[section]
\newtheorem{claim}{Claim}[section]
\newtheorem{definition}{Definition}[section]
\newtheorem{example}{Example}[section]
\newenvironment{notation}{\noindent\bf Notation:\em\penalty1000}{}
\newcommand{\blackslug}{\mbox{\hskip 1pt \vrule width 4pt height 8pt 
depth 1.5pt \hskip 1pt}}
\newcommand{\QED}{\quad\blackslug\lower 8.5pt\null\par\noindent}
\newcommand{\proof}{\par\penalty-1000\vskip .5 pt\noindent{\bf Proof\/: }}
\newcommand{\ru}{\rule[-0.4mm]{.1mm}{3mm}}
\newcommand{\nni}{\ru\hspace{-3.5pt}}
\newcommand{\sni}{\ru\hspace{-1pt}}
\newcommand{\pre}{\hspace{0.28em}}
\newcommand{\post}{\hspace{0.1em}}
\newcommand{\NIm}{\pre\nni\sim}
\newcommand{\NI}{\mbox{$\: \nni\sim$}}
\newcommand{\notNIm}{\pre\nni\not\sim}
\newcommand{\notNI}{\mbox{ $\nni\not\sim$ }}
\newcommand{\NIVm}{\pre\nni\sim_V}
\newcommand{\NIV}{\mbox{ $\nni\sim_V$ }}
\newcommand{\notNIVm}{\pre\sni{\not\sim}_V\post}
\newcommand{\notNIV}{\mbox{ $\sni{\not\sim}_V$ }}
\newcommand{\NIWm}{\pre\nni\sim_W}
\newcommand{\NIW}{\mbox{ $\nni\sim_W$ }}
\newcommand{\NIWp}{\mbox{ $\nni\sim_{W'}$ }}
\newcommand{\notNIWm}{\pre\sni{\not\sim}_W\post}
\newcommand{\notNIW}{\mbox{ $\sni{\not\sim}_W$ }}
\newcommand{\eem}{\hspace{0.8mm}\rule[-1mm]{.1mm}{4mm}\hspace{-4pt}}
\newcommand{\EM}{\eem\equiv}
\newcommand{\notEM}{\eem\not\equiv}
\newcommand{\R}{\cal R}
\newcommand{\notR}{\not {\hspace{-1.5mm}{\cal R}}}
\newcommand{\bK}{{\bf K}}
\newcommand{\bKp}{${\bf K}^p$}
\newcommand{\oK}{$\overline {\bf K}$}
\newcommand{\bP}{{\bf P}}
\newcommand{\ga}{\mbox{$\alpha$}}
\newcommand{\gb}{\mbox{$\beta$}}
\newcommand{\gc}{\mbox{$\gamma$}}
\newcommand{\gd}{\mbox{$\delta$}}
\newcommand{\gep}{\mbox{$\varepsilon$}}
\newcommand{\gf}{\mbox{$\zeta$}}
\newcommand{\cA}{\mbox{${\cal A}$}}
\newcommand{\cB}{\mbox{${\cal B}$}}
\newcommand{\cC}{\mbox{${\cal C}$}}
\newcommand{\cE}{\mbox{${\cal E}$}}
\newcommand{\cF}{\mbox{${\cal F}$}}
\newcommand{\cK}{\mbox{${\cal K}$}}
\newcommand{\cL}{\mbox{${\cal L}$}}
\newcommand{\cM}{\mbox{${\cal M}$}}
\newcommand{\cR}{\mbox{${\cal R}$}}
\newcommand{\cU}{\mbox{${\cal U}$}}
\newcommand{\ab}{\mbox{\ga \NI \gb}}
\newcommand{\cd}{\mbox{\gc \NI \gd}}
\newcommand{\ef}{\mbox{\gep \NI \gf}}
\newcommand{\xe}{\mbox{$\xi$ \NI $\eta$}}  
\newcommand{\pht}{\mbox{$\varphi$ \NI $\theta$}}
\newcommand{\rt}{\mbox{$\rho$ \NI $\tau$}}
\newcommand{\Cn}{\mbox{${\cal C}n$}}
\newcommand{\CF}{\mbox{${\cal C}_{\cal F}$}}
\newcommand{\CW}{\mbox{${\cal C}_{W}$}}
\newcommand{\Pf}{\mbox{${\cal P}_{f}$}}
\newcommand{\leC}{\mbox{${\preceq_{\cC}}$}}
\newcommand{\leF}{\mbox{${\preceq_{\cF}}$}}
\newcommand{\lC}{\mbox{${\prec_{\cC}}$}}
\newcommand{\notlC}{\mbox{$\not \! \! \lC$}}
\newcommand{\lF}{\mbox{${\prec_{\cF}}$}}
\newcommand{\Z}{\mbox{$Z_{\cF}$}}
\newcommand{\ra}{\rightarrow}
\newcommand{\Ra}{\Rightarrow}
\newcommand{\eqdef}{\stackrel{\rm def}{=}}
\newcommand{\absv}[1]{\mid #1 \mid}
\newcommand{\vstar}{\mbox{$V\sstar_{\infty}$}}
\newcommand{\sumstar}{\mbox{$\sum\sstar$}}
\newcommand{\tilh}{\mbox{$\tilde{h}$}}
\newcommand{\tilep}{\mbox{$\tilde{\varepsilon}$}}
\newcommand{\tilf}{\mbox{$\tilde{f}$}}
\newcommand{\gahat}{\mbox{$\hat{\ga}$}}
\newcommand{\gafalse}{\mbox{$\ga\NI{\bf false}$}}
\newcommand{\sstar}{^{*}}
\newcommand{\calR}{\mbox{${\cal R}\sstar$}}
\newcommand{\subseteqf}{\mbox{$\subseteq_{f}$}}

\title{Nonmonotonic inference operations
\thanks{
This work was 
partially supported by a grant from the Basic Research
Foundation, Israel Academy of Sciences and Humanities and
by the Jean and Helene Alfassa fund for 
research in Artificial Intelligence. Its final write-up was performed while
the second author visited the Laboratoire d'Informatique Th\'{e}orique
et de Programmation, Universit\'{e} Paris 6.}}
\author{Michael Freund \\ D\'{e}partement de Math\'{e}matiques, \\
Universit\'{e} d'Orl\'{e}ans, 45067 Orl\'{e}ans, C\'{e}dex 2 (France) \and
Daniel Lehmann \\ Department of Computer Science, \\
Hebrew University, Jerusalem 91904 (Israel)
}
\date{}
\maketitle
\begin{abstract}
A.~Tarski~\cite{Tar:56} proposed the study of infinitary consequence operations
as the central topic of mathematical logic.
He considered monotonicity to be a property of all such operations.
In this paper, we weaken the monotonicity requirement and consider more 
general operations, inference operations.
These operations describe the nonmonotonic logics both humans and machines
seem to be using when infering defeasible information from incomplete
knowledge.
We single out a number of interesting families of inference operations.
This study of infinitary inference operations is inspired by the results
of~\cite{KLMAI:89} on finitary nonmonotonic operations, but this paper
is self-contained.
\end{abstract}

\section{Introduction}
\label{sec:intro}
Since Tarski~\cite{Tar:56} and Gentzen~\cite{Gent:69}, 
logicians have had to choose between two possible frameworks for the study 
of logics. 
A set \cL\  of formulas being given, one may model a logic by a mapping 
\mbox{$\cC : 2^{\,\cL} \longrightarrow 2^{\,\cL}$} and 
for every subset $X$ of \cL, $\cC(X)$ is understood to be the set of 
all the consequences of the set $X$ of assumptions. 
But one may think that only finite sets of assumptions have well-defined
consequences, in which case one generally represents a logic by a relation 
$\vdash$ between 
a finite set of formulas on the left and a single formula on the right.
For a finite set $A$ of formulas and a formula $b$, the relation 
\mbox{$A \vdash b$} holds when $b$ is a consequence of the finite set $A$ 
of assumptions.
It turns out that, in the study of mathematical logic, the mappings \cC\ 
of interest are monotonic, i.e., satisfy 
\mbox{${\cal C}(X) \subseteq {\cal C}(Y)$} whenever \mbox{$X \subseteq Y$}.
Very often they are also compact, i.e., whenever \mbox{$a \in {\cal C}(X)$}, 
there exists a finite subset $A$ of $X$ such that \mbox{$a \in {\cal C}(A)$}. 
For compact monotonic mappings, the two frameworks are equivalent 
since, given an infinitary consequence operation \cC\ one may define a 
finitary relation $\vdash$ by:
\[ A \vdash a \Leftrightarrow a \in {\cal C}(A) \] and, given a finitary 
relation $\vdash$, one may define an operation \cC\ by:
\[\begin{array}{lll}
{\cal C}(X) & = & \{ a \mid {\rm there \ exists \ a \ finite \ set\ } A \subseteq X \\
& & {\rm \ such \ that \ } A \vdash a \}
\end{array}\]
and this provides a bijection between the two formalisms.
The choice of a framework for a study of nonmonotonic logics has more 
serious consequences, since, even for compact mappings, 
there is no bijection as in the monotonic case.
At this point, we do not know whether there is a reasonable notion 
of {\em pseudo-compactness} for which any finitary operation has 
a unique {\em pseudo-compact} extension.
It is only recently that the general study of nonmonotonic logics 
has begun and both frameworks have been used.
So far, \cite{Gabbay:85}, \cite{KLMAI:89},\cite{Leh:89}, \cite{LMTARK:90}, 
and \cite{LMAI:92} opted for the finitary framework, 
whereas \cite{Mak:89} and \cite{Mak:90} opted for the infinitary framework.
A different comparison of those two approaches may be found in~\cite{DM:92}. 
The present work will build a bridge between the two frameworks.
More precisely, since the finitary framework has, so far, 
been richer in technical results and provided more insight 
in defining interesting families of relations, 
this paper will extend to the infinitary framework most of the notions 
and results obtained in the finitary one.
Mainly, three types of question will be studied.
First, we shall try to define families of infinitary inference operations 
that are similar to the finitary families defined in \cite{KLMAI:89}.
In doing so, we shall notice some unsuspected complications 
stemming from the fact that finitary properties that are equivalent sometimes 
have infinitary analogues that are not.      
Then, we shall deal with the problem of extending finitary operations to 
infinitary ones. This is a crucial question in deciding which framework
to prefer. If there are finitary operations that cannot be extended, then
the study of infinitary operations will be of little use for learning 
about finitary ones.
It will be shown that, for each one of 
the families of interest, every finitary operation has a suitable extension, 
but, in general, more than one such extension.
In fact, we shall even propose a canonical way of extending operations 
that maps (almost) each finitary family into its analogue.
The third type of questions we shall deal with is the obtention of 
representation theorems that sharpen the representation theorems 
of \cite{KLMAI:89} and \cite{LMAI:92}. 

Another aspect of this work must be clarified here.
One may consider the underlying language of formulas \cL\ in a number of 
possible ways.
One may suppose nothing about it, as did D.~Makinson in his first efforts,
and, more recently, D.~Lehmann in~\cite{Leh:91}.
But many interesting families of operations may be defined only if one 
assumes that \cL\ comes with a monotonic consequence operation and
is closed under certain connectives.
One may assume that \cL\ is a classical propositional language as was 
done in~\cite{KLMAI:89}.
But, although this is probably agreeable to the Artificial Intelligence 
community, this seems to go against the main trend in Logics, 
where classical logic is only one of many possible logics.
Therefore, we take here the view that there is some underlying 
monotonic logic, on top of which we build a nonmonotonic inference operation, 
but will assume as little as possible about this underlying logic.
When specific assumptions about the existence of well-behaved connectives
will be necessary, they will be made explicitly.

\section{Plan of this paper}
\label{sec:plan}
This paper describes and studies a number of families of inference
operations, roughly from the most general to the most restricted.
First, in Section~\ref{sec:back}, we present the background we need concerning
monotonic consequence operations, and inference operations.
Basic definitions and results about connectives have been relegated
to Appendix~\ref{app:cL}.
Then, five principal families of inference operations are presented.

In Section~\ref{sec:cum}, we consider the {\em cumulative} operations. 
Cumulative operations correspond, in our more abstract setting, to what 
D.~Makinson called supraclassical 
cumulative inference operations in \cite{Mak:89}. 
The finitary version of this definition characterizes, in the setting of 
classical propositional calculus, the cumulative relations 
studied in~\cite{KLMAI:89}.  
The main results of this section deal with the problem of extending finitary 
cumulative inference relations into infinitary cumulative operations. 
Two such extensions are described. 
The first one, due to D.~Makinson, is the smallest possible cumulative 
extension of a given cumulative relation. 
The second one, termed {\em canonical} extension, is the generalization of
a construction presented in~\cite{FLM:89}.
We conclude this section by model-theoretic considerations: 
as in the finitary case, any  cumulative operation may be defined 
by a suitable model.
This representation theorem is a relatively straightforward sharpening of the
corresponding result of~\cite{KLMAI:89}.

Section~\ref{sec:loop} deals with {\em strongly cumulative} operations.
In the setting of classical propositional calculus, strongly cumulative
finitary operations are those cumulative operations that satisfy
the Loop property defined in \cite{KLMAI:89}.
The first of the extensions described in Section~\ref{sec:cum} extends 
a finitary cumulative operation satisfying Loop into an infinitary strongly 
cumulative operation. 
We do not know whether this is the case for the canonical extension.
We then show that, similarly to the finitary case, the strongly cumulative 
operations are precisely those that may be represented by a transitive model.

Section~\ref{sec:dist} defines {\em distributive} operations.
Distributivity is the property that corresponds, in the infinitary framework
and our more abstract setting, to the Or rule of~\cite{KLMAI:89}.
A weak form of distributivity is shown to be often equivalent to 
distributivity.
Properties of distributivity are studied. 
We find that the notion of a distributive finitary operation requires the
existence of a disjunction in the language, and we show that the canonical
extension of a distributive finitary operation is distributive.
The lack of a representation theorem for distributive operations seems 
to indicate this family is not as well-behaved as others. 

In Section~\ref{sec:ded}, another family of cumulative operations,
the {\em deductive} operations, is defined. 
They are the operations that satisfy an infinitary version of the S rule 
of~\cite{KLMAI:89}. 
We show that, although the rules Or and S are equivalent
for finitary cumulative relations, their infinitary counterparts
are not, since
there are, even in the setting of classical propositional calculus, 
distributive operations that are not deductive. 
We, then, study in detail the relations between distributive and deductive 
operations.
We show that the canonical extension of a deductive finitary operation is 
deductive.
We characterize monotonic deductive operations as translations 
of the operation of logical consequence.  
We, then, study Poole systems, generalize some previous results of
D.~Makinson and present some new ones.
We show, in particular, that any finite Poole system without constraints 
defines a deductive operation that is the canonical extension of its finitary 
restriction.
A representation theorem for deductive operations is established, 
that is a non-obvious generalization of the corresponding result 
of~\cite{KLMAI:89}.

In Section~\ref{sec:rat}, we deal with an important family of 
deductive operations, 
those that satisfy the infinitary version of Rational Monotonicity. 
We show that the canonical extension of a rational relation provides a 
rational operation and conclude with a representation theorem for rational 
operations in terms of modular models.
Section~\ref{sec:conc} is a conclusion.

\section{Background}
\label{sec:back}
Our treatment of nonmonotonic inference operations supposes some underlying
monotonic logic is given and inference operations behave reasonably with
respect to this logic.
On one hand, we would like our treatment of nonmonotonic inference
operations to be as general as possible and not to be tied to any
specific language or underlying monotonic logic. 
Therefore, we take an abstract view of the underlying logic and
make only minimal assumptions about it.
On the other hand, certain specific properties of the underlying logic, 
for example, the existence of a well-behaved disjunction or implication,
are sometimes needed to prove interesting results.
We shall define those properties in an Appendix and mention them explicitly 
when they shall be needed.

We suppose a non-empty set \cL\ of formulas is given.
Our canonical example for such a set is the set of all propositional formulas
on a given non-empty set of propositional variables and this is the set \cL\ 
we shall use in all our examples. 
But we do not assume anything about \cL\ in general.
Together with \cL\ we suppose there is a basic logic, syntax, semantics
and model theory attached. 
We assume that some compact consequence operation $\Cn$, 
in the sense of Tarski, is given. The operation $\Cn$ represents our notion of
{\em logical consequence}.
We shall often say \cL\ when we mean the couple 
\mbox{$\langle {\cal L} , \Cn \rangle$}, or even \cL, $\Cn$ and their model 
theory.
We shall use the symbol $\subseteqf$ to mean {\em is a finite subset of}.
We summarize now the properties we assume for \Cn.
They hold for arbitrary \mbox{$X , Y \subseteq {\cal L}$}.
\[
\begin{array}{ll}
({\rm Inclusion}) & X \subseteq \Cn(X) \\
({\rm Monotonicity}) & X \subseteq Y \Ra \Cn(X) \subseteq \Cn(Y) \\
({\rm Idempotence}) & \Cn(\Cn(X)) = \Cn(X) \\
({\rm Compactness}) & {\rm if} a \in \Cn(X) , \\
& {\rm there \  is \  a \  finite \  subset \ } A \subseteqf X \\ 
& {\rm such \  that \ } a \in \Cn(A)
\end{array}
\]
As is usually done, for \mbox{$ X , Y \subseteq \cL$}
and \mbox{$a \in \cL$}, we shall write 
\mbox{$\Cn(X , Y)$} instead of \mbox{$\Cn(X \cup Y)$},
\mbox{$\Cn(X ,a)$} instead of \mbox{$\Cn(X \cup \{ a \})$}
and \mbox{$\Cn(a)$} instead of \mbox{$\Cn(\{ a \})$}.
\begin{definition}
\label{def:theory}
A set of formulas $T$ that is closed under \Cn, i.e., such that
\mbox{$T = \Cn(T)$} is called a theory.
\end{definition}
We recall the fact that the intersection of a family of theories is a theory:
\begin{lemma}
\label{le:intertheories}
Let \mbox{$X_{i} , i \in I$} be a family of theories.
Their intersection \mbox{$\bigcap_{i \in I} X_{i}$} 
is a theory.
\end{lemma}
\proof
Let \mbox{$Y \eqdef \bigcap_{i \in I} X_{i}$}.
By Inclusion we have \mbox{$Y \subseteq \Cn(Y)$}.
By Monotonicity, we have \mbox{$\Cn (Y) \subseteq \Cn (X_{i}) = X_{i}$}
for \mbox{$i \in I$}.
Therefore:
\mbox{$ \Cn (Y) \subseteq \bigcap_{i \in I} X_{i} = Y$}.
\QED
For \mbox{$X , Y \subseteq \cL$} we shall say that $X$ is {\em consistent} 
iff \mbox{$\Cn(X) \neq \cL$} and that $X$ is {\em consistent with} 
$Y$ iff \mbox{$\Cn(X , Y) \neq \cL$}. A set is {\em inconsistent} iff it
is not consistent.

So far, our discussion has been purely syntactical and proof-theoretic.
We shall also suppose that, with the language \cL\ and the
consequence operation \Cn\ comes a suitable semantics in the form
of a set \cU\ (the universe), the elements of which we shall call worlds,
and a relation $\models$ of satisfaction between worlds and formulas.
We assume that, for any \mbox{$X \subseteq \cL$} and any \mbox{$a \in \cL$}:
\mbox{$a \in \Cn(X)$} iff any world \mbox{$w \in \cU$} that satisfies
all the formulas of $X$ also satisfies $a$.

A number of properties of the language \cL\ and the operation \Cn\ will be
used in the sequel. The definition of those properties: existence of 
connectives, the notion of a characteristic formula and admissibility
have been relegated to an Appendix.

We shall now introduce our definitions and notations for inference operations.
Let $S$ be a set, $2^{S}$ denotes the set of all subsets of $S$
and $\Pf(S)$ denotes the set of all finite subsets of $S$.
We shall consider infinitary operations (in short operations)
\mbox{$\cC : 2^{\, \cL} \longrightarrow 2^{\, \cL}$}
and finitary operations \mbox{$\cF : \Pf(\cL) \longrightarrow 2^{\, \cL}$}.
\begin{definition}
\label{def:CCn}
\ 
\begin{enumerate}
\item
An operation \cC\ is {\em left absorbing} iff,
for any \mbox{$X \subseteq \cL$}, $\cC(X)$ is a theory.
A finitary operation \cF\ is said to be {\em left absorbing} iff,
for any finite \mbox{$A \subseteqf \cL$}, $\cF(A)$ is a theory.
\item
An operation \cC\ is {\em right absorbing} iff,
for any \mbox{$X , Y \subseteq \cL$} such that \mbox{$\Cn(X) = \Cn(Y)$},
we have \mbox{$\cC(X) = \cC(Y)$}, or, equivalently, if
for any \mbox{$X \subseteq \cL$}, \mbox{$\cC(\Cn(X)) = \cC(X)$}.
A finitary operation \cF\ is {\em right absorbing} iff,
for any finite \mbox{$A , B \subseteqf \cL$} such that 
\mbox{$\Cn(A) = \Cn(B)$}, \mbox{$\cF(A) = \cF(B)$}.
\end{enumerate}
\end{definition}
\begin{definition}
\label{def:infop}
An (resp.\ a finitary) operation that is left-absorbing,
right-absorbing and satisfies Inclusion (i.e., \mbox{$X \subseteq \cC(X)$}) 
(resp.\ \mbox{$A \subseteqf \cF(A)$})
is called an (resp.\ a finitary) {\em inference operation}.
\end{definition}
Notice that any operation \cC\ (resp.\ finitary operation \cF) that
satisfies Inclusion and is left-absorbing is supraclassical, i.e.,
for any set of formulas
$X$, \mbox{$\Cn(X) \subseteq \cC(X)$} (resp.\ for any finite set of formulas $A$,
\mbox{$\Cn(A) \subseteq \cF(A)$}).
We shall use, for inference operations, the same notations we use for
consequence operations and write, for example \mbox{$\cC(X , Y)$}
instead of \mbox{$\cC(X \cup Y)$}.
Finitary inference operations are the analogue, in our more general
(as far as \cL\ and \Cn\ are concerned) framework, of those nonmonotonic
consequence relations of~\cite{KLMAI:89} that satisfy Reflexivity, Left Logical
Equivalence, Right Weakening and And.
Inference (both finitary and infinitary) operations represent ways
of infering defeasible conclusions from incomplete information.

Left absorption means that the logical consequences of 
the {\em adventurous} inferences made by \cC\ are already themselves
obtainable by \cC.
An inference operation is right absorbing if its inferences do not depend
on the form of the assumptions but only on their logical meaning.
As remarked in~\cite{Mak:90}, absorption properties seem to be necessary
characteristics of {\em logical} as opposed to {\em procedural} approaches.

Right absorption means that the image of any set of formulas is a theory.
Left absorption means that the image of any set $X$ is the same as that of
its associated theory $\Cn(X)$. We may, therefore, consider an inference
operation as an arbitrary mapping of theories into theories that satisfies
Inclusion, i.e., for any theory $T$, \mbox{$T \subseteq \cC(T)$}.
We shall do so freely.
Another definition will be handy.
\begin{definition}
\label{def:C-cons}
Let \cC\ be an inference operation and $X$ a set of formulas. 
We shall say that $X$ is \cC-consistent iff \mbox{$\cC(X) \neq \cL$}.
\end{definition}

\section{Cumulative operations}
\label{sec:cum}      
\subsection{Definition and first properties of cumulative operations}
\label{subsec:defcum}
We shall define a family of infinitary inference operations, the family
of cumulative operations, 
and show that this family is the exact analogue
of the family of finitary cumulative operations that has been studied
in~\cite{KLMAI:89} under the name of cumulative nonmonotonic consequence
relations.
Historically, the study of cumulative operations was launched independently
in two separate efforts.
D.~Makinson defined and studied cumulative infinitary operations 
with a consequence operation \Cn\ equal to the identity. 
S.~Kraus and D.~Lehmann defined and studied cumulative finitary operations, 
in the framework of classical propositional calculus.
This paper presents a generalization of Makinson's approach that 
benefits from the insights gained by studying finitary operations.
The terminology is mainly Makinson's.
\begin{definition}
\label{def:cuminf}   
An operation \cC\ is {\em cumulative} 
iff it is an inference operation and satisfies the following two properties.
For any \mbox{$X , Y \subseteq {\cal L}$} such that 
\mbox{$Y \subseteq \cC(X)$}:
\[
\begin{array}{ll}
({\rm Cut}) & \cC(X , Y) \subseteq \cC(X) \\
({\rm Cautious\ Monotonicity}) & \cC(X) \subseteq \cC(X , Y).
\end{array}
\]
A finitary operation \cF\ is {\em cumulative}
iff it is a finitary inference operation and satisfies, for all finite
\mbox{$A , B \subseteq {\cal L}$} such that 
\mbox{$B \subseteq \cF(A)$}:
\begin{tabbing}
{\rm (Finitary Cut)} $\qquad\quad\qquad$ \= $\cF(A , B) \subseteq \cF(A)$ \\
{\rm (Finitary Cautious Monotonicity)} \\
\> $\cF(A) \subseteq \cF(A , B)$.\\
\end{tabbing}
\end{definition}
The definition of infinitary (resp.\ finitary) cumulative
operations (for an inference operation) could have been given
in one go, by saying that, under the assumptions, 
\mbox{$\cC(X , Y) = \cC(X)$} (resp.\ \mbox{$\cF(A , B) = \cF(A)$}),
but we prefered to define the two
properties of Cut (resp.\ Finitary Cut) and Cautious Monotonicity
(resp.\ Finitary Cautious Monotonicity).
The following summarizes some easy results most of which appear 
in~\cite{Mak:89}.
\begin{theorem}
\label{the:summary}
Any (resp.\ finitary) operation that satisfies Supraclassicality, Cut
(resp.\ Finitary Cut) and Cautious Monotonicity (resp.\ Finitary Cautious
Monotonicity) is an inference operation, and therefore cumulative.
An (resp.\ a finitary) inference operation \cC\ (resp.\
\cF) is cumulative iff for all subsets \mbox{$X , Y \subseteq \cL$} such that
\mbox{$Y \subseteq \cC(X)$} and \mbox{$X \subseteq \cC(Y)$}
(resp.\ finite subsets \mbox{$A , B \subseteq \cL$} such that
\mbox{$B \subseteq \cF(A)$} and \mbox{$A \subseteq \cF(B)$})
one has
\mbox{$\cC(X) = \cC(Y)$}
(resp.\ \mbox{$\cF(A) = \cF(B)$}).
\end{theorem}
Supraclassicality expresses the fact that our inference procedures are
to be at least as adventurous as \Cn.
We need now to clear up the relation existing between the cumulative 
{\em finitary}
operations defined above and the cumulative 
{\em consequence relations} 
of~\cite{KLMAI:89}.
If we take \cL\ to be a propositional calculus and \Cn\ to be classical
logical consequence, then, the two notions coincide.
Given any finitary inference operation \cF, one may define a nonmonotonic
consequence relation \NI\ by: 
\mbox{$ a \NI b$} iff \mbox{$b \in \cF( a )$}.
It is easy to see that, if \cF\ is cumulative, so is \NI.
Given any nonmonotonic consequence relation \NI\ one may define a
finitary inference operation \cF\ by:
\mbox{$b \in \cF(A)$} iff \mbox{$\chi_{A} \NI b$}.
It is easy to see that, if \NI\ is cumulative, so is \cF,
and that, in this case, the two transformations above are inverse 
transformations.

We shall not try here to justify our interest in cumulative operations.
The reader may wish to consult~\cite{KLMAI:89} and~\cite{Mak:90} for 
motivation.
It is not clear, though, whether finitary or infinitary operations
are the topic of choice. We shall, therefore, devote, now, our efforts to
studying the relation existing between finitary and infinitary cumulative 
operations.
First, it is clear that, given any infinitary cumulative operation,
its restriction to finite sets provides a finitary cumulative operation.
A natural question is: may all finitary cumulative operations be obtained 
in this manner?  
In other words, given a finitary cumulative operation, can it be cumulatively 
extended to infinite sets.
This is an important question because a positive answer will enable
us to prove certain results for infinitary operations and then
apply them to finitary ones.
As an example, Theorem~\ref{the:cumrep} will provide us with a strengthening
of the completeness part of Theorem 3.25 of~\cite{KLMAI:89}.
The next sections will provide a positive answer to the question.
We shall propose two different ways of extending finitary cumulative
operations.
It follows that a finitary cumulative operation may have many different
cumulative extensions.

\subsection{The smallest cumulative extension of a finitary operation}
\label{subsec:Makext}
The first construction we shall present is due to D.~Makinson.
It requires no assumption on the language \cL.
It shows not only that every cumulative finitary operation may be
cumulatively extended but also that there is a smallest such extension,
smallest in the sense that the sets $\cC(X)$ are small.

Let \cF\ be a cumulative finitary operation.
We are looking for a cumulative operation \cC\ such that
\mbox{$\cC(A) = \cF(A)$} for any finite set $A$.
Let \mbox{$X \subseteq \cL$}.
Suppose that there is a finite set $A$ such that
\mbox{$A \subseteqf \Cn(X) \subseteq \cF(A)$}
(in such a case we shall say that $X$ is of the {\em first type}). 
Then, for any suitable \cC, one must have 
\mbox{$\Cn(X) \subseteq \cC(A)$} and, by cumulativity,
\mbox{$\cC(A , \Cn(X)) = \cC(A)$}.
But \mbox{$A \subseteq \Cn(X)$} and \mbox{$\cC(\Cn(X)) = \cC(X)$}
by right absorption. We conclude that
\mbox{$\cC(X) = \cC(A) = \cF(A)$}.
There is therefore only one possibility in defining \mbox{$\cC(X)$}
for an $X$ of the first type.

It turns out that, for such an $X$, one may define unambiguously 
\mbox{$\cC(X)$} to be $\cF(A)$ for any finite subset $A$ of $\Cn(X)$
such that \mbox{$X \subseteq \cF(A)$}.
Indeed, for any two finite sets \mbox{$A_{i} , i = 0 , 1$}, such that 
\mbox{$A_{i} \subseteq \Cn(X) \subseteq \cF(A_{i})$}, for \mbox{$i = 0 , 1$},
one has \mbox{$A_{0} \subseteq \cF(A_{1})$} and
\mbox{$A_{1} \subseteq \cF(A_{0})$}.
Since \cF\ is cumulative we have \mbox{$\cC(A_{0}) = \cC(A_{1})$}.
There is therefore exactly one way to define the extension of \cF\ on
sets of the first type.
We are left with a choice about the definition of $\cC(X)$ for $X$'s
that are not of the first type.
The {\em smallest} possible choice is to set: 
\mbox{$\cC(X) = \Cn(X)$} in such cases.
We shall show that the \cC\ obtained in this manner is cumulative and 
extends \cF.
It is clear that it is the smallest possible such operation.

\begin{theorem}
\label{the:makext}
For any cumulative finitary operation \cF, the operation
\cC\ defined by:
\[
\cC(X) = \left \{ 
	\begin{array}{ll}
		\cF(A)		& \mbox{if there is a finite subset $A$} \\
				& \mbox{of $\Cn(X)$ such that} \\
				& \Cn(X) \subseteq \cF(A), \\
				& \mbox{or equivalently, } X \subseteq \cF(A) \\
		\Cn(X)		& \mbox{otherwise}
	\end{array}
	\right.
\]
extends \cF and is cumulative.
\end{theorem}
\proof
Let $A$ be a finite set of formulas.
It is obviously of the first type since 
\mbox{$A \subseteqf \Cn(A) \subseteq \cF(A)$}.
By definition, then, we have \mbox{$\cC(A) = \cF(A)$}.
Let us show now that \cC\ is cumulative, and first that it is supraclassical.
Let X be any subset of \cL. 
We have to check that \mbox{$\Cn(X) \subseteq  \cC(X)$}. 
This is straightforward if $X$ is not of the first type.
Suppose it is of the first type.
There exists a finite set $A$ such that  
\mbox{$\Cn(X) \subseteq \cF(A)$} and \mbox{$\cC(X) = \cF(A)$}.
We have \mbox{$\Cn(X) \subseteq \cF(A) = \cC(X)$}.

Let us now show that \cC\ satisfies Cut and Cautious Monotonicity. 
We suppose that $X$ and $Y$ are subsets of \cL\  such that 
\mbox{$Y \subseteq  \cC(X)$}, and we want to prove that 
\mbox{$\cC(X , Y) = \cC(X)$}.
If $X$ is not of the first type, \mbox{$\cC(X) = Cn(X)$}, therefore 
\mbox{$Y \subseteq \Cn(X)$} and \mbox{$\Cn(X , Y) = \Cn(X)$}. 
We conclude that $X \cup Y$ is not of the first type either.
Hence  \mbox{$\cC(X , Y) = \Cn(X , Y) = \Cn(X) = \cC(X)$}.
If, on the contrary, $X$ is of the first type, there exists a finite subset 
\mbox{$A \subseteqf  \Cn(X)$} such that \mbox{$X \subseteq \cF(A) = \cC(X)$}.
We have
\mbox{$A \subseteqf \Cn(X , Y)$}
and
\mbox{$X \cup Y \subseteq \cF(A)$}, since, by hypothesis,
\mbox{$Y \subseteq \cC(X)$}.
Therefore \mbox{$\cC(X , Y) = \cF(A) = \cC(X)$}.
\QED
The smallest cumulative extension described above is a very elegant technical 
construction
but seems to fail to be a {\em natural} way of extending finitary operations.
For $X$'s that are not of the first type, there is no relation between
$\cC(X)$ and the $\cF(A)$'s for the finite subsets $A$ of $X$.
In particular, one may check that, even if the finitary \cF\ is monotonic,
its smallest cumulative extension is not in general monotonic.

For instance, take \cL\  to be the propositional calculus on the variables  
\mbox{$q , p_{0} , \ldots , p_{i} , \ldots$} and define \cF\ by: 
\mbox{$\cF(A) = \Cn(A , \{ q \})$}.
The finitary operation \cF\ is obviously cumulative and monotonic. 
Let $X$ be the set of all the $p_{i}$'s. 
Then it is readily seen that $X$ is not of the first type and, thus, that 
\mbox{$\cC(X) = \Cn(X)$}.
We conclude that \mbox{$q \in \cC(\{ p_{0} \})$} but
\mbox{$q \not\in \cC(X)$}.
   
We shall see further that the smallest cumulative extension does not preserve
other important properties, such as distributivity.
We shall present another method for extending finitary operation,
that provides more natural extensions, even though it is more convoluted.

\subsection{The canonical extension}
\label{subsec:can}
Let, again, \cF\ be a cumulative finitary operation. 
First, we shall define an infinitary inference operation \CF.
The definition of \CF\ is quite a natural variation on the idea of
compact extensions for monotonic finitary operations.
In short we shall introduce $a$ in $\CF(X)$ iff there is a finite
subset $A$ of $X$ such that, not only $a$ is in $\cF(A)$, but $a$ is 
also in $\cF(A , B)$ for any finite set \mbox{$B \subseteq \Cn(X)$}
(caution, we have $\Cn(X)$ not just $X$).
Intuitively, this means that $a$ will be inferred from $X$ iff $a$ is inferred 
from any finite subset of $\Cn(X)$ that is big enough.
Note that this is quite a conservative (small) extension.
One could perhaps consider some larger extensions.

K.~Schlechta~\cite{SchleInf:91} provided an example showing that the operation \CF\
is not always cumulative. This result will not be used in this paper.
We provide an iterative construction that starts at \CF\ and provides a cumulative extension
of any cumulative \cF.
For this, at some point, we shall need to assume that the language \cL\ has 
implication. The question of whether this requirement may be weakened,
for example to admissibility, is open, but the conjecture seems implausible.
When \cF\ satisfies some additional properties, to be studied in Section~\ref{sec:dist}
and further, we shall show that \CF\ itself is cumulative (and more).
The iterative construction developed here, in addition to allowing us to
build, for any finitary cumulative operation, a cumulative extension 
that is nicer than the smallest cumulative extension, is of
independent interest. It has been used extensively in~\cite{NIL:90,Freund:92}.
\begin{definition}
\label{def:CF}
Let \cF\ be a cumulative finitary operation.
The operation \CF\ is defined in the following way: 
for any \mbox{$X \subseteq \cL$},
\[\begin{array}{lll}
\CF(X) & \eqdef & \{ a \in \cL \mid \exists A \subseteqf X {\rm \ such\ that\ } \\
& & a \in \cF(A , B) , \ \forall B \subseteqf \Cn(X) \}.
\end{array}
\]
\end{definition}
We shall now proceed to study the operation \CF.
We shall show that it is an inference operation and provides an almost 
cumulative extension.
Our first lemma generalizes the definition of \CF\ from formulas
to finite sets of formulas. It is a technical result needed in the sequel.
\begin{lemma}
\label{le:tech}
Let \mbox{$X \subseteq \cL$} and $A$ a finite subset of $\CF(X)$.
There is a finite subset \mbox{$B \subseteqf X$} such that
\mbox{$A \subseteqf \cF(B , C)$} for any finite \mbox{$C \subseteqf \Cn(X)$}.
\end{lemma}
\proof
Let $a$ be an arbitrary element of $A$.
Since $a$ is an element of $\CF(X)$, there is a finite subset $B_{a}$ of $X$ 
such that \mbox{$a \in \cF(B_{a} , C)$} for any finite subset $C$ 
of $\Cn(X)$.
Let $B$ be the union of all the $B_{a}$'s, for \mbox{$a \in A$}.
It is a finite subset of $X$.
For any finite \mbox{$C \subseteqf \Cn(X)$}, and for any \mbox{$a \in A$},
\mbox{$B_{a} \subseteq B \cup C \subseteqf \Cn(X)$} and therefore
\mbox{$a \in \CF(B , C)$}.
\QED
\begin{lemma}
\label{le:ext}
The operation \CF\ is an extension of \cF, i.e., for any finite set $C$,
\mbox{$\CF(C) = \cF(C)$}.
\end{lemma}
\proof
Suppose $C$ is finite.
Let, first, $a$ be an element of $\CF(C)$. We shall show it is an element
of $\cF(C)$.
Indeed, let $A \subseteq C$ be as in Definition~\ref{def:CF}.
Since $C$ is a finite subset of $\Cn(C)$, 
\mbox{$a \in \cF(A , C) = \cF(C)$}.

Let, now, $a$ be an element of $\cF(C)$. We shall show it is an element
of $\CF(C)$.
Take the $A$ of Definition~\ref{def:CF} to be $C$.
For any finite subset $B$ of $\Cn(C)$, we have
\mbox{$\cF(C) = \cF(C , B)$} by right-absorption of \cF.
Therefore \mbox{$a \in \cF(C , B)$}.
We conclude that $a$ is in $\CF(C)$.
\QED
\begin{lemma}
\label{le:supra}
The operation \CF\ is supraclassical.
\end{lemma}
\proof
Let $X$ be a subset of \cL, and $a$ an element of $\Cn(X)$. 
We must show that $a$ is in $\CF(X)$. 
But, since \Cn\ is compact, there is a finite subset $A$ of $X$ such that 
\mbox{$a \in \Cn(A)$}. 
Let $B$ be any finite set.
We have \mbox{$a \in \Cn(A , B)$} and, since \cF\ is supraclassical,
\mbox{$a \in \cF(A , B)$}.
We conclude from Definition~\ref{def:CF} that \mbox{$a \in \CF(X)$}.
\QED
\begin{lemma}
\label{le:cut}
The operation \CF\ satisfies Cut.
\end{lemma}
\proof   
Suppose \mbox{$Y \subseteq \CF(X)$}. 
We want to show that \mbox{$\CF(X , Y) \subseteq  \CF(X)$}.
Let $a$ be an element of the first set. 
There exists a finite subset $A$ of $X \cup Y$ such that $a$ is in 
\mbox{$\cF(A , B)$} for any finite subset $B$ of $\Cn(X , Y)$. 
Note that $A$ is a finite subset of $\CF(X)$. 
It follows, then, from Lemma~\ref{le:tech} that there is a finite subset 
$A'$ of $X$ such that \mbox{$A \subseteq \cF(A' , B)$} 
for any finite subset $B$ of $\Cn(X)$.
To show that \mbox{$a \in \CF(X)$}, we shall take the $A$ of 
Definition~\ref{def:CF} to be $A'$.
Let $B$ be any finite subset of $\Cn(X)$.
We must show that \mbox{$a \in \cF(A' , B)$}.
But we know that \mbox{$A \subseteq \cF(A' , B)$}.
Since \cF\ satisfies finitary Cut, we conclude that
\mbox{$\cF(A' , B , A) \subseteq \cF(A' , B)$}.
But \mbox{$A' \cup B \subseteq \Cn(X) \subseteq \Cn(X , Y)$}
and therefore
\mbox{$a \in \cF(A , A' , B)$}.
We conclude that \mbox{$a \in \cF(A' , B)$}.
\QED
\begin{lemma}
\label{le:absorb}
The operation \CF\ satisfies both left and right absorption,
and is therefore an inference operation.
\end{lemma}
\proof
For left absorption, one easily checks that any supraclassical operation \cC\
that satisfies Cut, satisfies left absorption.
Indeed \mbox{$\Cn(\cC(X) \subseteq$} 
\nolinebreak[3]  \mbox{$\cC(\cC(X))$} by supraclassicality.
But \mbox{$\cC(X) \subseteq$} \nolinebreak[3] $\cC(X)$ and, by Cut, we have
\mbox{$\cC(X , \cC(X)) \subseteq$} \nolinebreak[3] $\cC(X)$ and therefore
\mbox{$\cC(\cC(X)) \subseteq$} \nolinebreak[3] $\cC(X)$.
For right absorption, one first shows that any supraclassical operation \cC\
that satisfies Cut, satisfies \mbox{$\cC(\Cn(X)) \subseteq$} \nolinebreak[3]
$\cC(X)$.
Indeed \mbox{$\Cn(X) \subseteq \cC(X)$} by supraclassicality.
By Cut, we conclude that \mbox{$\cC(X , \Cn(X)) \subseteq \cC(X)$}.
We now need to show that
\mbox{$\CF(X) \subseteq \CF(\Cn(X))$}.
Let $a$ be an element of the first set. 
Let \mbox{$A \subseteqf X$} be the finite set promised by 
Definition~\ref{def:CF}.
It is a finite subset of $\Cn(X)$, and, since $a$ is an element
of \mbox{$\cF(A , B)$} for any finite \mbox{$B \subseteq \Cn(X)$},
it is so for any finite \mbox{$B \subseteqf \Cn(\Cn(X))$}.
\QED
As alluded to before, \CF\ does not satisfy Cautious Monotonicity. 
Nevertheless, it satisfies some special cases of it.
\begin{definition}
\label{def:spcm}
\ 
\begin{enumerate}
\item \label{1-spcm}
An operation \cC\ is said to satisfy 1-special Cautious Monotonicity 
iff, for any \mbox{$Y \subseteq \cL$} and any {\em finite} 
\mbox{$A \subseteqf \cL$}, \mbox{$Y \subseteq  \cC(A)$} implies  
\mbox{$\cC(A) \subseteq  \cC(A , Y)$}.
\item \label{2-spcm}
An operation \cC\ is said to satisfy 2-special Cautious Monotonicity 
iff, for any \mbox{$X \subseteq \cL$} and any {\em finite} 
\mbox{$B \subseteqf \cL$}, \mbox{$B \subseteq  \cC(X)$} implies  
\mbox{$\cC(X) \subseteq  \cC(X , B)$}.
\end{enumerate}
\end{definition}
It is easy to check that, in part~\ref{2-spcm} of this definition, one could 
have considered only the case where $B$ is a singleton.
Similarly, we can define special cases of Cut.
\begin{definition}
\label{def:spc}
\ 
\begin{enumerate}
\item
An operation \cC\ is said to satisfy 1-special Cut iff, 
for any \mbox{$Y \subseteq \cL$} and any {\em finite} 
\mbox{$A \subseteqf \cL$}, \mbox{$Y \subseteq  \cC(A)$} implies  
\mbox{$\cC(A , Y) \subseteq  \cC(A)$}.
\item
An operation \cC\ is said to satisfy 2-special Cut iff, 
for any \mbox{$X \subseteq \cL$} and any {\em finite} 
\mbox{$B \subseteqf \cL$}, \mbox{$B \subseteq  \cC(X)$} implies  
\mbox{$\cC(X , B) \subseteq  \cC(X)$}.
\end{enumerate}
\end{definition}
\begin{lemma}
\label{le:1spcm}
The operation \CF\ satisfies 1-special Cautious Monotonicity.
\end{lemma}
\proof
Suppose that $Y$ is a subset of $\CF(A)$.
We want to show that \mbox{$\CF(A) \subseteq  \CF(A , Y)$}.
Note, first, that \mbox{$\CF(A) = \cF(A)$} by Lemma~\ref{le:ext}. 
Let $a$ be an arbitrary element of $\cF(A)$.
The set $A$ is a finite subset of \mbox{$A \cup Y$}. We shall take it as the 
$A$ of Definition~\ref{def:CF} and show that 
\mbox{$a \in \cF(A , B)$} for any finite subset $B$ of
\mbox{$\Cn(A , Y)$}. 
Let $B$ be such a finite set. 
Since \mbox{$B \subseteq  \cF(A)$}, we see, by finitary cautious monotonicity
that \mbox{$\cF(A) \subseteq  \cF(A , B)$}.
Therefore \mbox{$a \in \cF(A , B)$}.
\QED
For the next lemma, some assumption on the language \cL\ is needed.
However, one may notice that, if \Cn\ is the identity, the result holds
without any assumption on the language.
\begin{lemma}
\label{le:2spcm}
If the language \cL\ has implication, 
then the operation \CF\ 
satisfies 2-special Cautious Monotonicity.
\end{lemma}
\proof
Suppose that $B$ is a finite subset of $\CF(X)$.
We want to show that \mbox{$\CF(X) \subseteq  \CF(X , B)$}.
By Lemma~\ref{le:tech}, there is a finite subset $A'$ of $X$ such that 
\mbox{$B \subseteq \cF(A' , C)$} for any finite subset $C$ of $\Cn(X)$. 
Let $a$ be an arbitrary element of $\CF(X)$. 
We must prove that \mbox{$a \in \CF(X , B)$}.
By Definition~\ref{def:CF} there is a finite subset $A$ of $X$, 
such that $a$ is an element of \mbox{$\cF(A , C)$} for any finite subset 
$C$ of $\Cn(X)$. 
The set \mbox{$A \cup A'$} is a finite subset of \mbox{$X \cup B$}.
This is the set we shall use as the $A$ of Definition~\ref{def:CF}.
We must show that \mbox{$a \in \cF(A , A' , C)$} for any finite
\mbox{$C \subseteqf \Cn(X , B)$}.
Let $C$ be such a set. 
By part~\ref{bconimp1} of Lemma~\ref{le:basic} we know that: 
\mbox{$B \ra C \subseteqf \Cn(X)$}.
From the defining property of $A$, we see that
\mbox{$a \in \cF(A , A' , B \ra C)$}.
It is left to us to show that
\mbox{$\cF(A , A' , B \ra C) \subseteq \cF(A , A' , C)$}.
But, by the defining property of $A'$, we see that
\mbox{$B \subseteq \cF(A , A' , B \ra C)$} and therefore we have
\mbox{$C \subseteq \cF(A , A' , B \ra C)$}.
By finitary cautious monotonicity we conclude that 
\mbox{$\cF(A , A' , B \ra C) \subseteq \cF(A , A' , B \ra C , C)$}.
But \mbox{$B \ra C \subseteq \Cn(C)$} and by, right absorption,
\mbox{$\cF(A , A' , B \ra C , C) = \cF(A , A' , C)$} and we are through.
\QED
Our goal is now to transform the extension \CF\ in an iterative way,
to obtain an extension that is cumulative.
The following general transformation will prove interesting.
\begin{definition}
\label{def:trans}
Let \cC\ be any operation.
We shall define its transform $\cC'$ by:
\[\begin{array}{lll}
\cC'(X) & = & \{ a \in \cL \mid \exists A \subseteqf X {\rm \ such\ that\ } \\
& & a \in \cC(A , Y) , \ \forall Y \subseteq \cC(X) \}.
\end{array}
\]
\end{definition}
A fuller study of the properties of this transformation may be found 
in~\cite{NIL:90} and~\cite{Freund:92}.
Our first remark, that will not be used, is that the transform of any 
operation is compact (take $Y$ to be empty).
The next lemma is a technical lemma, its proof follows exactly the line
of that of Lemma~\ref{le:tech} and will be omitted.
\begin{lemma}
\label{le:tech2}
Let \cC\ be any operation that satisfies Inclusion.
Let \mbox{$X \subseteq \cL$} and $A$ a finite subset of $\cC'(X)$.
There is a finite subset \mbox{$B \subseteqf X$} such that
\mbox{$A \subseteqf \cC(B , C)$} for any finite \mbox{$C \subseteqf \cC(X)$}.
\end{lemma}
We shall now study the properties of the transformation we just defined.
\begin{lemma}
\label{le:Inc}
If \cC\ satisfies Inclusion, so does its transform $\cC'$.
\end{lemma}
\proof
Suppose \mbox{$a \in X$}.
Take the $A$ of Definition~\ref{def:trans} to be the singleton $\{a\}$.
\QED
\begin{lemma}
\label{le:smaller}
Let \cC\ be any operation that satisfies Inclusion.
Its transform $\cC'$ is smaller than \cC, i.e., 
for any $X$, 
\mbox{$\cC'(X) \subseteq \cC(X)$}.
\end{lemma}
\proof
Let $a$ be an element of $\cC'(X)$.
Let $A$ be the finite set promised by Definition~\ref{def:trans}.
Since \mbox{$X \subseteq \cC(X)$}, we conclude that
\mbox{$a \in \cC(A , X) = \cC(X)$}.
\QED
\begin{lemma}
\label{le:supra'}
If \cC\ is supraclassical, then so is its transform $\cC'$.
\end{lemma} 
\proof
Let \cC\ be a supraclassical operation.
We want to show that, for any $X$,  \mbox{$\Cn(X) \subseteq  \cC'(X)$}. 
Let $a$ be an arbitrary element of $\Cn(X)$.  
Since \Cn\ is compact, there exists a finite subset $A$ of $X$ such that 
\mbox{$a \in \Cn(A)$}. We shall choose this finite set as the $A$ of 
Definition~\ref{def:trans}.
But \mbox{$a \in \Cn(A , Y) \subseteq \cC(A , Y)$} for any set $Y$.
We conclude that $a$ is an element of $\cC'(X)$ as desired.
\QED
\begin{lemma}
\label{le:ext'}
If \cC\ satisfies Inclusion and 1-special Cautious Monotonicity,
then \cC\ and $\cC'$ agree on finite sets.
\end{lemma}
\proof
Let \cC\ be an operation that satisfies 
1-special Cautious Monotonicity. 
Let $A$ be a finite subset of \cL. 
We must prove that \mbox{$\cC(A) = \cC'(A)$}. 
By Lemma~\ref{le:smaller}, it is enough to show that 
\mbox{$\cC(A) \subseteq  \cC'(A)$}.
Let $a$ be an element of the first set.
We shall take $A$ itself to be the $A$ of Definition~\ref{def:trans}.
Let $Y$ be an arbitrary subset of $\cC(A)$. 
We shall show that \mbox{$a \in \cC(A , Y)$}.
Indeed, since \cC\ satisfies 1-special Cautious Monotonicity, we deduce
from \mbox{$Y \subseteq \cC(A)$} that we have 
\mbox{$\cC(A) \subseteq  \cC(A , Y)$}.
\QED
The next lemma explains our interest in the transform of an operation.
It is a way of building operations that satisfy Cautious Monotonicity.
\begin{lemma}
\label{le:CutCM}
Let \cC\ be an operation that satisfies Inclusion.
\begin{enumerate}
\item if \cC\ satisfies Cut, then its transform satisfies 
Cautious Monotonicity.
\item  if \cC\ satisfies 1-special Cut, 
then its transform satisfies 1-special Cautious Monotonicity.
\item if \cC\ satisfies 2-special Cut, 
then its transform satisfies 2-special Cautious Monotonicity.
\end{enumerate}
\end{lemma}
\proof
For the first item, let \cC\ be an operation that satisfies Inclusion
and Cut. 
Let \mbox{$Y \subseteq \cC'(X)$}.
We must show that \mbox{$\cC'(X) \subseteq \cC'(X , Y)$}.
Let $a$ be an arbitrary element of $\cC'(X)$. 
By Definition~\ref{def:trans}, there exists a finite subset $A$ of $X$ 
such that \mbox{$a \in \cC(A , Z)$} for any subset $Z$ of $\cC(X)$.
But $A$ is a finite subset of \mbox{$X \cup Y$}, and we shall take it
to be the $A$ of Definition~\ref{def:trans} to show that $a$ is an element
of \mbox{$\cC'(X , Y)$}.
Take $Z$ to be an arbitrary subset of \mbox{$\cC(X , Y)$}.
We must show that \mbox{$a \in \cC(A , Z)$}.
But, since \cC\ satisfies Inclusion, by Lemma~\ref{le:smaller}, 
\mbox{$\cC'(X) \subseteq \cC(X)$},
and therefore \mbox{$Y \subseteq \cC(X)$}.
Since \cC\ satisfies Cut, now, we conclude that
\mbox{$\cC(X , Y) \subseteq \cC(X)$}.
Therefore $Z$ is a subset of $\cC(X)$ and \mbox{$a \in \cC(A , Z)$}.
For the last two items, the proofs are exactly the same, except that
certain sets are supposed to be finite.
\QED
\begin{lemma}
\label{le:CMCut}
Let \cC\ be an operation that satisfies Inclusion and 2-special Cut.
\begin{enumerate}
\item if \cC\ satisfies Cautious Monotonicity, 
then its transform satisfies Cut.
\item if \cC\ satisfies 1-special Cautious Monotonicity, 
then its transform satisfies 1-special Cut.
\item if \cC\ satisfies 2-special Cautious Monotonicity, 
then its transform satisfies 2-special Cut.
\end{enumerate}
\end{lemma}
\proof
We shall prove the first item. The other items are proved in exactly the 
same way, noticing that certain sets are finite.
Suppose \cC\ satisfies Inclusion, 2-special Cut and Cautious Monotonicity.
Suppose \mbox{$Y \subseteq \cC'(X)$}.
We want to show that \mbox{$\cC'(X , Y) \subseteq \cC'(X)$}.
Let $a$ be an arbitrary element of the first set.
By Definition~\ref{def:trans} there is a finite set 
\mbox{$A \subseteq X \cup Y$} such that \mbox{$a \in \cC(A , Z)$}
for any \mbox{$Z \subseteq \cC(X , Y)$}.
But, by Lemma~\ref{le:Inc}, $\cC'$ satisfies Inclusion and, by hypothesis,
\mbox{$Y \subseteq \cC'(X)$}. 
We conclude that \mbox{$A \subseteq \cC'(X)$}.
By Lemma~\ref{le:tech2}, there is a finite subset \mbox{$B \subseteq X$},
such that \mbox{$A \subseteq \cC(B , Z)$} for any 
\mbox{$Z \subseteq \cC(X)$}.
We claim that $B$ may be taken as the $A$ of Definition~\ref{def:trans}
and shall show that \mbox{$a \in \cC'(X)$} by showing that
\mbox{$a \in \cC(B , Z)$} for any \mbox{$Z \subseteq \cC(X)$}.
Take an arbitrary such $Z$.
We know that \mbox{$A \subseteq \cC(B , Z)$}.
Since $A$ is finite and \cC\ satisfies 2-special Cut,
\mbox{$\cC(B , Z , A) \subseteq \cC(B , Z)$}.
But, by Lemma~\ref{le:smaller}, \mbox{$\cC'(X) \subseteq \cC(X)$}
and therefore \mbox{$Y \subseteq \cC(X)$}.
Since \cC\ satisfies Cautious Monotonicity, we have
\mbox{$\cC(X) \subseteq \cC(X , Y)$}.
Therefore $Z$ is a subset of this last set.
Since \cC\ satisfies Inclusion, \mbox{$B \cup Z \subseteq \cC(X , Y)$}.
We conclude that \mbox{$a \in \cC(A , B , Z)$}.
Therefore, \mbox{$a \in \cC(B , Z)$}.
\QED
We are now ready to define the canonical extension of a finitary
cumulative operation.
\begin{definition}
\label{def:realcan}
Let \cF\ be any finitary operation and \CF\ the operation defined, out of
\cF, as described at the beginning of Section~\ref{subsec:can}.
Let us define $\cC_{0}$ to be \CF.
For any natural number $i > 0$, let us define $\cC_{i}$ to be the
transform of $\cC_{i-1}$.
The canonical extension of \cF\ is the intersection of the $\cC_{i}$'s,
i.e., the operation \cC\ such that, for any \mbox{$X \subseteq L$},
\mbox{$\cC(X) = \bigcap_{i \in \omega} \cC_{i}(X)$}.
\end{definition}
One may criticize our decision to call the operation defined in 
Definition~\ref{def:realcan} the canonical extension, since it is not
always equal to its transform.
It would have perhaps been wiser to call canonical extension the
operation obtained, by ordinal induction, when iterating the process of
taking the transform until we get some operation that is equal to its 
transform. All claims made below about about the canonical extension hold
true for this operation too: our definition uses, as {\em the} canonical extension,
the first (i.e. the largest) suitable operation found during our construction,
i.e. the closest to \CF.
\begin{theorem}
\label{the:canext}
If the language \cL\ has implication, 
the canonical extension of any 
cumulative finitary operation is indeed an extension and is cumulative.
\end{theorem}
\proof
Lemmas~\ref{le:ext}, \ref{le:supra}, \ref{le:cut}, \ref{le:1spcm} 
and~\ref{le:2spcm} show that \CF\ is a supraclassical extension of \cF\
that satisfies Cut and the two forms of special Cautious Monotonicity.
Lemmas~\ref{le:Inc}, \ref{le:CutCM} and \ref{le:CMCut} imply that
all $C_{i}$'s satisfy Inclusion, the even ones satisfy Cut and both forms
of special Cautious Monotonicity, and the odd ones satisfy 
Cautious Monotonicity and both special forms of Cut.
Lemma~\ref{le:supra'} implies that all $C_{i}$'s are supraclassical.
Lemmas~\ref{le:ext} and \ref{le:ext'} now imply that all $C_{i}$'s agree with 
\cF\ on finite sets.
We conclude that the canonical extension of a cumulative finitary operation
is indeed an extension (i.e., agrees with the original operation on 
finite sets) and is supraclassical.
By Lemma~\ref{le:smaller}, the chain of the $C_{i}$'s is a descending chain,
therefore the canonical extension is the intersection of the even $C_{i}$'s.
They all satisfy Cut, and an intersection of operations satisfying Cut
satisfies Cut. We conclude that the canonical extension satisfies Cut.
Similarly it satisfies Cautious Monotonicity (consider the odd indexes).
\QED
We may remark that the canonical extension of a cumulative finitary
operation is compact: indeed, all $C_{i}$'s are compact and they all agree on 
finite sets.

As will be seen in Section~\ref{sec:dist}, many finitary operations
\cF, have a \CF\ that is equal to its transform.
In such a case, obviously, \CF\ is the canonical extension.
We may remark that this is also the case if \cF\ is any monotonic operation
(not even cumulative). 
Then its \CF\ is its compact extension and is therefore monotonic too.
Clearly, in this case, the operation \CF\ is equal to its transform and is the 
canonical extension of \cF.
We have, so far, provided two different ways of extending finitary 
cumulative operations. In the next section we deal with the model-theory
of cumulative operations.

\subsection{Cumulative models}
\label{subsec:cummod}
We shall now present a natural way of defining cumulative operations.
We shall define a family of models and describe the operation defined
by a model. We shall show that all such models define cumulative operations and
that all such operations may be defined this way.
In~\cite{KLMAI:89}, this was done for finitary cumulative operations. 
The results presented here provide a sharpening of those of~\cite{KLMAI:89}.
The methods used are very similar.
The reader should consult~\cite{KLMAI:89} for motivation and background.

We recall first, some definitions concerning binary relations.
If $R$ is a binary relation, \mbox{$a \notR b$} will mean that the pair
\mbox{$(a , b)$} does not stand in the relation $R$.
\begin{definition}
\label{def:prelimord}
Let $\prec$ be a binary relation on a set $U$ and let \mbox{$V \subseteq U$}. 
We shall say that
\begin{enumerate}
\item \label{asy}
$\prec$ is {\em asymmetric} iff 
\mbox{$\forall s , t \in U$} such that 
\mbox{$s \prec t$}, we have
\mbox{$t \not \prec s$},
\item
\label{minal}
\mbox{$t \in V$} 
is {\em minimal} in $V$ iff
\mbox{$\forall s\in V$},
\mbox{$s \not \prec t$},
\item \label{minum}
\mbox{$t \in V$} 
is a {\em minimum} of $V$ iff \/
\mbox{$\forall s\in V$} such that \mbox{$s \neq t$},
\mbox{$t \prec s$},
\item \label{smoo}
$V$ is  {\em smooth} iff
\mbox{$\forall t \in V$}, either 
\mbox{$\exists s$} minimal in $V$, such that \mbox{$s \prec t$} or
$t$ is itself minimal in $V$.
\end{enumerate}
\end{definition}
We shall use the following lemmas, the proofs of which are obvious.
\begin{lemma}
\label{min:min}
Let $U$ be a set and $\prec$ an asymmetric binary relation on 
$U$. If\ $U$ has a minimum it is unique, it is a minimal element of $U$ and
$U$ is smooth.
\end{lemma}

\begin{definition}
\label{def:cum.mod}
A {\em cumulative} model is a triple 
\mbox{$\langle \: S , l , \prec \rangle$},
where $S$ is a set, the elements of which are called states, 
\mbox{$l:S \mapsto 2^{\,\cU}$} is a function that labels every state 
with a non-empty set of worlds and $\prec$ is a binary relation on $S$, 
satisfying the {\bf smoothness condition} that will be defined below
in Definition~\ref{def:smoocond}.
\end{definition}
The notion of a {\em world} used here has been described in 
Section~\ref{sec:back}. Our worlds correspond to models of the underlying
calculus, or sets of formulas. Our states correspond to what are often
called worlds in the framework of Kripke semantics.
Notice that the relation $\prec$ is an arbitrary binary relation.
It represents the preference one may have between different states,
or the degree of normality of a state, i.e., if \mbox{$s \prec t$},
$s$ is preferred to or more typical than $t$.
\begin{definition}
\label{def:EM}
Let \mbox{$\langle \: S , l , \prec \rangle$} be as above.
If \mbox{$X \subseteq \cL$} is a set of formulas, 
we shall say that $s \in S$ satisfies $X$ and write
\mbox{$s \EM X$} iff for every world \mbox{$m \in l(s)$}, and every
formula \mbox{$a \in X$}, 
\mbox{$m \models a$}.
The set:
\mbox{$\{ s \mid s \in S, \: s \EM X \}$} 
of all states that satisfy $X$ will be denoted by
\mbox{$\widehat{X}$}.
\end{definition}
\begin{definition}[smoothness condition]
\label{def:smoocond}
A triple
\mbox{$\langle \: S , l , \prec \rangle$} is said to satisfy
the smoothness condition iff, for any set \mbox{$X \subseteq \cL$}
of formulas, the set \mbox{$\widehat{X}$} is smooth.
\end{definition}
The reader must be cautioned that the definition given here to cumulative
models is more restrictive than the one given in~\cite{KLMAI:89}.
The smoothness property required here is stronger than the one
presented there. In~\cite{KLMAI:89}, only sets of the form $\widehat{A}$
for {\em finite} sets $A$ of formulas were required to be smooth. 
If we need to refer to the models of~\cite{KLMAI:89}, we shall refer to them
as finitary cumulative models.
Any cumulative model is finitary cumulative. The converse does not hold.
We shall now describe how a cumulative model defines an inference operation.
\begin{definition}
\label{def:cumcons}
Suppose a cumulative model \mbox{$W = \langle S , l, \prec \rangle$} is given. 
The inference operation defined by $W$ will be denoted by $\cC_{W}$ 
and is defined by:
\[\begin{array}{lll}
\cC_{W}(X) & = & \{ a \in \cL \mid s \EM a \\
& & {\rm \ for\ every\ } s {\rm \ minimal\ in\ }
\widehat{X} \}
\end{array}
\]
\end{definition}
\begin{theorem}
\label{the:cumsou}
Let $W$ be a cumulative model.
The operation $\cC_{W}$ is a cumulative operation.
\end{theorem}
\proof
Let us show, first, that \CW\ is supraclassical.
Let $X$ be a set of formulas.
All the minimal states of $\widehat{X}$ satisfy $X$ and therefore satisfy
$\Cn(X)$.

We shall show, now, that Cut is satisfied. 
Suppose that \mbox{$Y \subseteq  \CW(X)$}.
We want to show that \mbox{$\CW(X , Y) \subseteq  \CW(X)$}. 
Let $a$ be an element of the first set and $s$ a state minimal in 
$\widehat{X}$. 
We have to check that $s$ satisfies $a$. 
Note that $s$ satisfies $\CW(X)$ by the very definition of \CW. 
It follows that $s$ satisfies $Y$, and \mbox{$X \cup Y$}.
It is therefore an element of $\widehat{X \cup Y}$.
Now, we claim that $s$ is minimal in $\widehat{X \cup Y}$,
since it is minimal in the larger set $\widehat{X}$.
Therefore $s$ satisfies \mbox{$\CW(X , Y)$} and in particular, $a$.

For Cautious Monotonicity, suppose that $Y \subseteq  \CW(X)$.
We want to show that $\CW(X) \subseteq  \CW(X , Y)$. 
Let $a$ be an element of the first set and $s$ a state minimal in 
$\widehat{X \cup Y}$. We have to check that $s$ satisfies $a$. 
Note that $s$ satisfies $X$ and lies in $\widehat{X}$.
We shall show that $s$ is minimal in $\widehat{X}$.
Suppose, therefore, $s$ is not minimal in $\widehat{X}$.
By the smoothness property, there is a state $t \prec s$ such that
$t$ is minimal in $\widehat{X}$. But $t$ satisfies $\CW(X)$ and hence
$Y$. It is therefore an element of \mbox{$\widehat{X \cup Y}$}.
But $s$ is minimal in this set. A contradiction.
We have shown that $s$ is minimal in $\widehat{X}$. Therefore $s$ satisfies 
$\CW(X)$ and, in particular, $a$.
\QED
We are now interested in the converse problem: 
given a cumulative operation \cC, does there exist a cumulative 
model $W$ such that  \mbox{$\cC = \CW$}?  
As in the finitary case, treated in~\cite{KLMAI:89}, the answer is positive.
Suppose an arbitrary cumulative operation \cC\ is given.
We shall, first, define the set of states $S$ of our model.
Two subsets $X$ and $Y$ of \cL\ will be said  to be \cC-equivalent 
iff \mbox{$\cC(X) = \cC(Y)$}. 
This defines an equivalence relation on the subsets of \cL. 
Note that $X$ and $Y$ are logically equivalent iff they are \Cn-equivalent. 
Let us denote by $[X]$ the \cC-equivalence class of a set $X$.
We put $S$ to be the set of all such equivalence classes and
define the relation $\prec$ among the elements of $S$ by:
\[\begin{array}{ll}
[X] \prec [Y] & {\rm iff\ } [X] \neq [Y] {\rm \ and\ } \exists X' \subseteq \cC(Y) \\
& {\rm such\ that\ } X' {\rm \ is\ } \cC\mbox{\rm -equivalent\ to\ } X.
\end{array}
\]    
This is a well-defined binary relation among the elements of $S$,
i.e., the definition does not depend on the representatives used.
Let us note immediately the following.
\begin{lemma}
\label{le:asy}
The relation $\prec$ is asymmetric.
\end{lemma}
\proof
Suppose we have \mbox{$[X] \prec [Y]$} and \mbox{$[Y] \prec [X]$}. 
We shall derive a contradiction. 
Then, there are two sets $X'$ and $Y'$ such that  
\mbox{$\cC(X') = \cC(X)$}, \mbox{$\cC(Y') = \cC(Y)$}, $X'$ is a subset of 
$\cC(Y)$ and $Y'$ a subset of $\cC(X)$. 
Now, this implies that $X'$ is a subset of $\cC(Y')$ and $Y'$ a subset of 
$\cC(X')$. 
Since \cC\ is cumulative, we  have \mbox{$\cC(X') = \cC(Y')$}, 
hence \mbox{$\cC(X) = \cC(Y)$}, which contradicts 
\mbox{$[X] \prec [Y]$}.
\QED
To complete the definition of the model $W$, we define the function $l$ by
(remember \cU\ is the universe, defined in Section~\ref{sec:back}) 
\[
l([X]) = \{ m \in \cU \mid m \models \cC(X) \}. 
\]
One checks easily this does not depend on the choice of
the representative of [X].
\begin{lemma}
\label{le:X}
A state $[X]$ is an element of $\widehat{Y}$ iff $Y$ is a subset of $\cC(X)$.
\end{lemma}
\proof  
Indeed $[X]$ is an element of $\widehat{Y}$ iff $l([X])$ satisfies $Y$. 
By the way $l$ was defined, this holds iff any world $m$ that satisfies 
\cC(X), satisfies $Y$.
This is equivalent to \mbox{$Y \subseteq \Cn(\cC(X))$}. 
We conclude by left absorption.
\QED
\begin{corollary}
\label{co:1}  
The state $[X]$ is the minimum of $\widehat{X}$.
\end{corollary}
\proof
From Lemma~\ref{le:X} and the definition of $\prec$. 
\QED
\begin{corollary}
\label{co:2}  
The state $[X]$ is minimal in $\widehat{X}$ and $\widehat{X}$ is smooth.
\end{corollary}
\proof
Straightforward, using corollary~\ref{co:1} and Lemmas~\ref{le:asy}
and \ref{min:min}. 
\QED
\begin{theorem}
\label{the:cumrep}
Any cumulative operation may be defined by a cumulative model
in which the preference relation is an asymmetric relation such that, 
for every set $X$ of formulas $\widehat{X}$ has a minimum.
\end{theorem}
\proof:      
Let indeed \cC\ be a cumulative operation and 
\mbox{$W = \langle S , l , \prec \rangle$} defined as above. 
By corollary~\ref{co:2}, $W$ is a cumulative model. 
Let us check that \mbox{$\CW = \cC$}.
A formula $a$ is in $\CW(X)$ iff it is satisfied by all minimal states of 
$\widehat{X}$. 
By Lemma~\ref{le:X} and corollary~\ref{co:1} this happens iff $a$ is satisfied 
by $[X]$, hence iff any world $m$ that satisfies $\cC(X)$ satisfies also $a$.
But this is just equivalent to \mbox{$a \in \Cn(\cC(X)) = \cC(X)$}.
\QED
Notice that no assumption on the language \cL\ is made in 
Theorem~\ref{the:cumrep}.
By putting together our results about the extension of a finitary 
cumulative operation and Theorem~\ref{the:cumrep},
we may now conclude that any finitary cumulative operation may be defined
by a cumulative model, or more precisely is the restriction to finite sets
of the operation defined by some cumulative model.
In other words, in~\cite{KLMAI:89}, finitary cumulative operations 
were shown to be representable by models satisfying a weak smoothness property,
we showed here they may be represented also by
models that enjoy the stronger smoothness property.
A completeness result, weaker than Theorem~\ref{the:cumrep}, appears
in~\cite{Mak:90}.

\section{Strong cumulativity}
\label{sec:loop}
\subsection{Introduction}
\label{subsec:loopintro}
The cumulative models described in Section~\ref{subsec:cummod} contain a 
preference relation that is not required to be a partial ordering.
It seems reasonable, though, that {\em preference} should be transitive.
We therefore study operations that satisfy an additional property, 
stronger than cumulativity, but weaker than monotonicity, which we call
{\em strong cumulativity}.
Strong cumulativity is the infinitary version of the Loop property 
of~\cite{KLMAI:89}. 
We show, then, that the smallest cumulative extension of a finitary operation
that satisfies Loop is a strongly cumulative operation. 
Whether this is also true for the canonical extension of such an operation is 
not known to us. 
We finish by a representation theorem in which we show that operations that 
satisfy strong cumulativity are exactly those that can be defined by a 
cumulative model whose preference relation is a partial order.

\subsection{Strong cumulativity and the Loop property}
\label{subsec:stloop}
\begin{definition}
\label{def:strong}
An operation \cC\ is {\em strongly cumulative}
iff it is supraclassical and satisfies the following property, that
should be understood for any natural number $n$ and any sets of formulas
\mbox{$X_{i} , i = 0 , \ldots , n-1$} and where addition is understood modulo
$n$: 

\bigskip

\noindent {\rm (Strong Cumulativity)}
\begin{eqnarray*}    
{\rm If \ }& X_{i} \subseteq \cC(X_{i+1}) , {\rm for\ \ } i = 0 , \ldots , n-1 , \\ 
{\rm then \ }& \cC(X_{i}) = \cC(X_{j}) , {\rm for \ } 0 \leq i , j < n .
\end{eqnarray*}
A finitary operation \cF\ is strongly cumulative iff
it is supraclassical and satisfies the property obtained from the one above 
by requiring the $X_{i}$'s to be finite and replacing \cC\ by \cF.
\end{definition}
Notice that any strongly cumulative operation is cumulative.
Indeed, Strong Cumulativity in the case $n = 2$ is immediately seen 
to be the property mentioned in Theorem~\ref{the:summary}.
One also sees immediately that any monotonic cumulative operation 
is strongly cumulative. It is also immediate that, if \cL\ is classical
propositional calculus then Finitary Strong Cumulativity is exactly the
Loop property of~\cite{KLMAI:89}. 
Since, in~\cite{KLMAI:89}, a finitary inference operation that is
cumulative but does not satisfy Loop was described, it follows from the 
existence of cumulative extensions shown in Section~\ref{sec:cum} that there 
are cumulative operations that are not strongly cumulative: the cumulative
extension of the finitary operation mentioned above, for example.
We would like to know whether any finitary operation that is strongly 
cumulative may be extended to a strongly cumulative infinitary operation.
\begin{theorem}
\label{the:smstrong}
The smallest cumulative extension of a strongly cumulative finitary operation 
is strongly cumulative.
\end{theorem}
\proof
Let \cF\ be a strongly cumulative finitary operation and \cC\ its smallest
cumulative extension.
We already know from Theorem~\ref{the:makext} that \cC\ is cumulative.
It is therefore supraclassical.
We shall show that it satisfies Strong Cumulativity.
Let \mbox{$X_{i} , i = 0 , \ldots , n-1$} be such that
\mbox{$X_{i} \subseteq \cC(X_{i+1}) , i = 0 , \ldots , n-1$}
where addition is understood modulo $n$.
We have to prove that all the $X_{i}$'s are \cC-equivalent. 
We shall proceed by induction on $n$. 
For \mbox{$n = 2$}, this holds by Theorem~\ref{the:summary},
since \cC\ is cumulative, by Theorem~\ref{the:makext}. 

For the general case, we distinguish two cases.  
Suppose, first, that each one of the $X_{i}$'s is of the first type,
i.e., for any $i$ there is a finite set $A_{i}$ such that
\mbox{$A_{i} \subseteq \Cn(X_{i}) \subseteq \cF(A_{i})$}.
But, for any $i$, \mbox{$\Cn(X_{i}) \subseteq \cC(X_{i+1})$} and
also \mbox{$\cC(X_{i}) = \cF(A_{i})$}.
We conclude that \mbox{$A_{i} \subseteq \cF(A_{i+1})$}.
Since \cF\ is strongly cumulative, we are done.  

Suppose now there is some $X_{k}$ that is not of the first type.
In this case \mbox{$\cC(X_{k}) = \Cn(X_{k})$}.
Therefore we have \mbox{$X_{k-1} \subseteq \Cn(X_{k})$}
and \mbox{$X_{k} \subseteq \cC(X_{k+1})$}.
We easily conclude that \mbox{$X_{k-1} \subseteq \cC(X_{k+1})$}.
We can now apply the induction hypothesis to the sequence of $n-1$ sets
obtained by removing the set $X_{k}$ from the original sequence.
We conclude that \mbox{$\cC(X_{i}) = \cC(X_{j})$} for all $i$, $j$ different 
from $k$.
In particular we have \mbox{$\cC(X_{k+1}) = \cC(X_{k-1})$}.  
But \mbox{$X_{k} \subseteq \cC(X_{k+1})$}, and therefore 
\mbox{$X_{k} \subseteq \cC(X_{k-1})$}.
Since \mbox{$X_{k-1} \subseteq \cC(X_{k})$}, we conclude,
by the cumulativity of \cC, that \mbox{$\cC(X_{k}) = \cC(X_{k-1})$}
and this completes the proof.
\QED

\subsection{The representation of strongly cumulative operations}
\label{subsec:stcumrep}
We shall see now that the strongly cumulative operations are exactly those 
that can be defined by a cumulative model in which the preference relation 
$\prec$ is a strict partial order, i.e., is irreflexive and transitive. 
Let us call such models cumulative ordered models. 
\begin{theorem}
\label{the:soundcumord}
Let \mbox{$W = \langle S , l , \prec \rangle$} be a cumulative ordered model. 
Then the operation \CW\ induced by this model is strongly cumulative.
\end{theorem}
\proof
We only have to prove that \CW\ satisfies the property of Strong Cumulativity. 
Let $n$ be any natural number.
Suppose that, for \mbox{$0 \leq i \leq n-1$},
\mbox{$X_{i} \subseteq  \CW(X_{i+1})$} where addition is understood modulo $n$.
We shall prove that all the $X_{i}$'s are \CW-equivalent.
Let $s_{n-1}$ be any state minimal in $\widehat{X_{n-1}}$. 
Then $s_{n-1}$ satisfies $\CW(X_{n-1})$, and satisfies therefore $X_{n-2}$. 
By the smoothness property, there exists a state $s_{n-2}$ minimal in $X_{n-2}$
such that \mbox{$s_{n-2} \prec s_{n-1}$} or \mbox{$s_{n-2} = s_{n-1}$}. 
Repeating this argument leads to a sequence $s_{i}$ such that for  
\mbox{$0 \leq i \leq n-1$}, \mbox{$s_{i} \prec s_{i+1}$} or 
\mbox{$s_{i} = s_{i+1}$} where addition is modulo $n$.
The preference relation $\prec$ being transitive and irreflexive, 
all the $s_{i}$'s must be
equal.
Any state minimal in some $\widehat{X_{i}}$ is minimal in all 
$\widehat{X_{i}}$'s.
All $X_{i}$'s are therefore \CW-equivalent.
\QED
\begin{theorem}
\label{the:repcumstr}
Any strongly cumulative operation may be defined by some cumulative ordered 
model in which, for every set $X$ of formulas $\widehat{X}$ has a minimum.
\end{theorem}
\proof
Let \cC\ be a strongly cumulative operation, and 
\mbox{$W = \langle S , l , \prec \rangle$} the cumulative model defining
\cC\ that was described in Section~\ref{subsec:cummod} just before
Theorem~\ref{the:cumrep}.
Define $\prec^+$ to be the transitive closure of $\prec$  and put 
\mbox{$W' = \langle S , l , \prec^+ \rangle$}. 
To prove our theorem, it is enough to check that $\prec^+$ is irreflexive,
that $W'$ satisfies the smoothness condition and that 
\mbox{${\cal C}_{W'} = \cC$}.
We shall prove, first, that $\prec^+$ is irreflexive.
Suppose \mbox{$[X] \prec^+ [X]$}. 
There must be a finite sequence of states 
$[X_{i}]$, for \mbox{$0 \leq i \leq n-1$}, 
such that \mbox{$[X_{i}] \prec [X_{i+1}]$} for any $i$, 
\mbox{$0 \leq i \leq n-2$} and \mbox{$[X_{0}] = [X] = [X_{n-1}]$}.
Since $\prec$ is ireflexive, $n > 1$.
From the way $\prec$ was defined, there are sets $Y_{i}$, such that,
for any $i$,
\mbox{$\cC(X_{i}) = \cC(Y_{i})$}, \mbox{$Y_{i} \subseteq \cC(X_{i+1})$}.
But \mbox{$\cC(Y_{n-1}) = \cC(X_{n-1}) = \cC(X_{0}) = \cC(Y_{0})$}.
Therefore the $Y_{i}$'s satisfy the premisses of Strong Cumulativity.
But \cC\ is strongly cumulative. 
Therefore all the $Y_{i}$'s are \cC-equivalent, and the same holds for 
the $X_{i}$'s. A contradiction to \mbox{$[X_{i}] \prec [X_{i+1}]$}.
We have proven that  $\prec^+$ is irreflexive. 

Let us show that $W'$ satisfies the smoothness condition.
Since $\prec^+$ is irreflexive and transitive, it is asymmetric.
Let $X$ be a set of formulas.
We know that $[X]$ is the minimum of $\widehat{X}$ in $W$.
It is obviously also a minimum of $\widehat{X}$ under $\prec^+$.
We conclude by Lemma~\ref{min:min}.

Given any set $X$ of formulas, in both $W$ and $W'$, $[X]$ is the unique
minimal element of $\widehat{X}$, we conclude that
\mbox{$\cC_{W'}(X) = \CW(X) = \cC(X)$}.
\QED

\section{Distributive operations}
\label{sec:dist}
\subsection{Definition and basic facts}
\label{subsec:basic}
In the present section we shall study a restricted family of strongly
cumulative operations.
The definition of this family needs the consideration of some
intricate interaction between the inference operation \cC\ and the
operation of logical consequence \Cn.
We shall be able to prove interesting results about this family but 
shall not be able to provide a representation theorem for it.
Also, though the notion of infinitary distributivity may be defined
without any assumptions on the language \cL, the corresponding notion of 
{\em finitary distributivity} requires \cL\ to have disjunction.
Distributive inference operations are those strongly cumulative operations
that satisfy an additional property, that is the infinitary analogue 
(in a loose sense for the moment) of the Or rule of~\cite{KLMAI:89}.
The results obtained in~\cite{KLMAI:89} for the finitary operations that
satisfy this rule suggested to D.~Makinson the study of its infinitary
version. First results, in the setting of classical propositional calculus,
appear in~\cite{Mak:89}.
\begin{definition}
\label{def:distr}
An inference operation \cC\ is {\em distributive} iff it is 
cumulative and satisfies the following property, for any sets $X$, $Y$, $Z$
of formulas:

\bigskip

\noindent {\rm (Distributivity)}
\begin{equation}
\label{eq:dis}
\cC(Z , X) \cap \cC(Z , Y) \subseteq  
\cC(Z , \Cn(X) \cap \Cn(Y))
\end{equation}
\end{definition}
First some comments.
We have not justified, yet, our claim that distributive operations are 
strongly cumulative. This will be done in Theorem~\ref{the:diststr}.
Since any cumulative operation satisfies right absorption,
we could as well have formulated the property of Distributivity as:
for any theories $X$, $Y$, $Z$,
\[
\cC(Z , X) \cap \cC(Z , Y) \subseteq \cC(Z , X \cap Y)
\]
or, if the language \cL\ has disjunction, in view of property~\ref{bdis} of
Lemma~\ref{le:basic}, as:
for any sets $X$, $Y$, $Z$,
\[
\cC(Z , X) \cap \cC(Z , Y) \subseteq \cC(Z , X \vee Y)
\]
This latest formulation is probably the most telling one:
if some formula may be inferred, in the presence of assumptions $Z$, 
both from $X$ and from $Y$,
then it may be infered from their disjunction, in the presence of $Z$.
There are supraclassical, monotonic operations that satisfy Cut but
do not satisfy Distributivity. Therefore, contrary to Cautious Monotonicity
and Strong Cumulativity, Distributivity is not a special case of Monotonicity.
Consider, for example, classical propositional calculus and define \cC\ by:
\[
\cC(X) = \left \{ 
	\begin{array}{ll}
		\cL		& \mbox{if $\Cn(X) \neq \Cn(\emptyset)$} \\
		\Cn(\emptyset) 	& \mbox{otherwise.}
	\end{array}
	\right.
\]
The operation \cC\ is monotonic and cumulative, but not distributive. 
It is worth noticing that operations of logical consequence \Cn\ that satisfy
the Tarski conditions listed in Section~\ref{sec:back} are not always 
distributive. 
In fact the language \cL\ is admissible iff \Cn\ is distributive. 
It should, therefore, come as no surprise that we shall, 
for most of our results about distributive operations, 
have to assume that \cL\ is admissible.
Let us, now, try to say, in some more precise terms, in what sense 
Distributivity is the infinitary analogue of the Or rule.
We would like it to mean that there is a natural notion of Distributivity
for finitary inference operations that is equivalent to the Or rule,
when the setting is classical propositional calculus.
Since, for finite sets of formulas $A$, $B$, the intersection
\mbox{$\Cn(A) \cap \Cn(B)$} is not in general finite or even logically 
equivalent to some finite set (it is if there is a disjunction in the language),
the natural notion of Distributivity for finitary operations is:
for any finite sets of formulas $A$, $B$, $C$

\bigskip

\noindent {\rm (Finitary Distributivity)}
\[
\cF(C , A) \cap \cF(C , B) \subseteq \cF(C , A \vee B).
\]
But this is well defined only if the language \cL\ has disjunction. 
Now, classical propositional calculus has disjunction and, in this setting
a finitary cumulative operation satisfies Finitary Distributivity iff
it satisfies the Or rule of~\cite{KLMAI:89}.
Notice, now, that, Distributivity 
looks very much like one half of the usual property defining disjunction. 
More precisely, supposing \cL\ has disjunction, it implies that this 
disjunction behaves as half of a {\em disjunction for} \cC, i.e.
\mbox{$\cC(X , a) \cap \cC(X , b) \subseteq \cC(X , a \vee b)$}.
Distributivity is a bit stronger than this property, though, since it
allows us to consider sets of formulas instead of single formulas.
We shall now present a most useful tool in the study of distributive
operations: any distributive operation defines an ordering on theories.
This ordering may be interpreted as saying a theory is at least as normal
as another one.

\subsection{The ordering defined by a distributive operation}
\label{subsec:distriborder}
We shall define a binary relation on theories, for any inference operation
\cC.
\begin{definition}
\label{def:ord1}
Let \cC\ be an inference operation.
Let $X , Y$ be theories.
We shall say that \mbox{$X \leC Y$} iff 
\mbox{$X \subseteq \cC(X \cap Y)$},
\end{definition}
This relation expresses that theory $X$ is
{\em more expected}, or less unusual than theory $Y$, since $X$ is expected
if the formulas that are common to both theories are true.
If the language \cL\ has disjunction, the meaning of \leC\ 
is intuitively clear, \mbox{$X \leC Y$} iff \mbox{$X \subseteq \cC(X \vee Y)$}
meaning that, on the premise that either $X$ or $Y$
holds, one infers that $X$ holds.
For finitary inference operations, we shall use the following definition,
that assumes the language \cL\ has disjunction.
\begin{definition}
\label{def:ford1}
Assume \cL\ has disjunction.
Let \cF\ be a finitary inference operation.
Let \mbox{$A , B \subseteqf \cL$}.
We shall say that \mbox{$A \leF B$} iff 
\mbox{$A \subseteq \cF(A \vee B)$},
\end{definition}
It is clear that, in definition~\ref{def:ford1}, $A$ could have been replaced
by $\Cn(A)$ and $B$ by $\Cn(B)$; the relation $\leF$ is really a
relation between finitely generated theories.
The following lemma expresses some basic properties of the relation
$\leC$ for a {\em cumulative} operation \cC.
\begin{lemma}
\label{le:hi1}
Let \cC\ be a cumulative operation.
Let \mbox{$X , Y$} be theories.
\begin{enumerate}
\item \label{refl} \mbox{$X \subseteq Y \Ra X \leC Y$}, 
in particular, $\leC$ is reflexive,
\item \label{eq} \mbox{$X \leC Y$} iff 
\mbox{$\cC(X) = \cC(X \cap Y)$},
\item \label{incon} if \mbox{$X \leC Y$} and $X$ is \cC-inconsistent,
then $Y$ is \cC-inconsistent,
\item \label{anti} \mbox{$X \leC Y$} and \mbox{$Y \leC X$} imply 
\mbox{$\cC(X) = \cC(Y)$},
\item \label{XCX} \mbox{$X \leC \cC(X)$} and 
\item \label{CXX} \mbox{$\cC(X) \leC X$}.
\end{enumerate}
\end{lemma}
\proof
Item~\ref{refl} is proved by Inclusion.
Notice that the hypothesis is \mbox{$X \subseteq Y$}, not the
weaker \mbox{$X \subseteq \cC(Y)$}. The stronger property obtained
by weakening the hypothesis is the property of weak distributivity
that will be defined in item~\ref{weakd} of Theorem~\ref{the:distweak}.
One direction of item~\ref{eq} is proved by Cumulativity, the other one
by Inclusion.
For item~\ref{incon}, by item~\ref{eq}, 
\mbox{$\cC(X) = \cC(X \cap Y)$}. But \mbox{$\cC(X) = \cL$}.
Therefore \mbox{$Y \subseteq \cC(X \cap Y)$}.
We conclude by Cumulativity.
Item~\ref{anti} follows from~\ref{eq}. Notice that the converse of~\ref{anti}
is not claimed to hold.
Items~\ref{XCX} and~\ref{CXX} are proved by Supraclassicality.
\QED
The finitary version of Lemma~\ref{le:hi1} is the following. 
The proof is similar.
\begin{lemma}
\label{le:fhi1}
Assume \cL\ has disjunction.
Let \cF\ be a finitary cumulative operation.
Let \mbox{$A , B$} be finitely generated theories.
\begin{enumerate}
\item \label{frefl} \mbox{$A \subseteq B \Ra A \leF B$}, 
in particular, $\leF$ is reflexive,
\item \label{feq} \mbox{$A \leF B$} iff 
\mbox{$\cF(A) = \cF(A \vee B)$},
\item \label{fincon} if \mbox{$A \leF B$} and $A$ is \cF-inconsistent,
then $B$ is \cF-inconsistent,
\item \label{fanti} \mbox{$A \leF B$} and \mbox{$B \leF A$} imply 
\mbox{$\cF(A) = \cF(B)$}.
\end{enumerate}
\end{lemma}

\subsection{Weak Distributivity}
\label{subsec:weakdistrib}
We shall now consider a property of weak distributivity that is implied by 
Distributivity and very often equivalent to it.
Our first result concerns two equivalent formulations of this property.
One of them looks very weak and their equivalence is perhaps surprising.
Our proof represents a generalization of and an improvement on a similar 
proof by D.~Makinson and K.~Schlechta for the finitary case in the setting
of classical propositional calculus. 
Notice that no assumption on \cL\ is needed.
We state and prove here the infinitary version of the result. 
The finitary version holds true and is proved in a completely similar way.
\begin{theorem}
\label{the:distweak}
Let \cC\ be a cumulative operation. The following two properties
are equivalent and satisfied by any distributive operation. 
\begin{enumerate}
\item \label{w1} For any theories $X$, $Y$, 

$\cC(X) \cap \cC(Y) \subseteq \cC(X \cap Y)$.
\item \label{weakd} For any sets of formulas $X$, $Y$,\\
if $Y \subseteq \cC(X)$, then $Y \leC X$.
\end{enumerate}
\end{theorem}
\proof
Property~\ref{w1} is a special case of Distributivity: the case when the $Z$ 
of Definition~\ref{def:distr} is empty.
Property~\ref{weakd} expresses the very natural 
(and at first sight weak) property
that if a set of formulas, $Y$, may be inferred from $X$, they may be inferred 
from \mbox{$X \vee Y$}.
Let us show that property~\ref{w1} implies property~\ref{weakd}.
Suppose \cC\ is cumulative and satisfies property~\ref{w1}.
If \mbox{$Y \subseteq \cC(X)$}, we have
\mbox{$Y \subseteq \cC(X) \cap \cC(Y)$} by Inclusion.
Since
\mbox{$\cC(X) \cap \cC(Y)$} \mbox{$\subseteq \cC(X \cap Y)$},
we are easily done.

The other direction is more delicate.
Suppose \cC\ is a cumulative operation that satisfies property~\ref{weakd}.
To show that it satisfies property~\ref{w1} we shall consider three arbitrary
{\em theories} $X$, $Y$ and $Z$ and show that if 
\mbox{$Z \subseteq$} \nolinebreak[3] \mbox{$\cC(X) \cap \cC(Y)$}, then 
\mbox{$Z \subseteq$} \nolinebreak[3] \mbox{$\cC(X \cap Y)$}.
Property~\ref{w1} will follow by taking \mbox{$Z =$} \nolinebreak[3]
\mbox{$\cC(X) \cap \cC(Y)$},
this last intersection being a theory.
First, since \mbox{$Z \subseteq$} \nolinebreak[3] \mbox{$\cC(X)$}, 
we notice that
\mbox{$\Cn(X , Z) \subseteq$} \nolinebreak[3] \mbox{$\cC(X)$}.
Let now $W$ be an arbitrary theory.
We have \mbox{$\Cn(X , Z) \cap W \subseteq$} \nolinebreak[3] \mbox{$\cC(X)$}.
We now apply property~\ref{weakd} and conclude that
\mbox{$\Cn(X , Z) \cap W \subseteq$} \nolinebreak[3]
\mbox{$\cC(X \cap \Cn(X , Z) \cap W) =$} \nolinebreak[3] 
\mbox{$\cC(X \cap W)$}.
We shall now take $W$ to be $Y$
and conclude that
\mbox{$\Cn( X , Z) \cap Y \subseteq$} \nolinebreak[3] \mbox{$\cC(X \cap Y)$}.
Since \cC\ satisfies Cut, we conclude that
\mbox{$\cC(X \cap Y , \Cn( X , Z) \cap Y) \subseteq$} \nolinebreak[3]
\mbox{$\cC(X \cap Y)$}.
Therefore we have
\mbox{$\cC(\Cn( X , Z) \cap Y) \subseteq$} \nolinebreak[3]
\mbox{$\cC(X \cap Y)$}.
It is left to us to prove that 
\mbox{$Z \subseteq$} \nolinebreak[3] \mbox{$\cC(\Cn( X , Z) \cap Y)$}.
But, similarly to what was done above, from the fact that
\mbox{$Z \subseteq$} \nolinebreak[3] \mbox{$\cC(Y)$}, we conclude that, 
for any theory $W$,
we have \mbox{$\Cn(Y , Z) \cap W \subseteq$} \nolinebreak[3]
\mbox{$\cC(Y \cap W)$}.
We shall take $W$ to be \mbox{$\Cn( X , Z)$} and are done.
\QED
Any inference operation that satisfies the properties of 
Theorem~\ref{the:distweak} will be said to be {\em weakly distributive}. 
The following shows that weak distributivity is not as weak as it seems
at first sight.
\begin{theorem}
\label{the:wf}
If \cL\ is admissible, any weakly distributive operation is distributive.
\end{theorem}
\proof
Suppose \cC\ satisfies property~\ref{w1} of Theorem~\ref{the:distweak}.
We have \linebreak[1]
\mbox{$\cC(Z , X) \cap \cC(Z , Y) \subseteq$} \nolinebreak[3]
\mbox{$\cC(\Cn(Z , X) \cap \Cn(Z , Y))$}.
But \cL\ is admissible, and
\mbox{$\Cn(Z , X) \cap \Cn(Z , Y) =$} \nolinebreak[3]
\mbox{$\Cn(Z , X \cap Y)$}.
One concludes by right-absorption.
\QED
The following will be useful. As usual, we state and prove only the infinitary
version here but the finitary version holds true and is proved similarly,
assuming \cL\ has disjunction.
\begin{lemma}
\label{le:usefdist}
Let \cC\ be a distributive operation. For any theories $X , Y , W , Z$:
\begin{enumerate}
\item \label{u1}
If \mbox{$Y \subseteq \cC(X)$}, then \mbox{$\cC(Y) = \cC(X \cap Y)$},
\item \label{u2}
if \mbox{$X \leC Y$} and \mbox{$W \leC Z$}, then 
\mbox{$X \cap W \leC Y \cap Z$}.
\end{enumerate}
\end{lemma}
\proof
For item~\ref{u1}, from the easy part of Theorem~\ref{the:distweak} 
we know that \mbox{$Y \leC X$}. We conclude by part~\ref{eq} of 
Lemma~\ref{le:hi1}.
For item~\ref{u2}, we have
\[
X \cap W \subseteq \cC(X \cap Y) \cap \cC(W \cap Z).
\]
We conclude by Distributivity.
\QED
The following result will be crucial in Section~\ref{sec:ded}.
\begin{theorem}
\label{the:order}
If the operation \cC\ is distributive, then the relation \leC\ is transitive,
and therefore a pre-order.
\end{theorem}
\proof
We have seen in part~\ref{refl} of Lemma~\ref{le:hi1} that \leC\ is reflexive.
Suppose \mbox{$X \leC Y$} and 
\mbox{$Y \leC Z$}.
We must show that \mbox{$X \leC Z$}.
Without loss of generality we may assume that $X$, $Y$ and $Z$ are theories.
From the hypotheses, by Lemma~\ref{le:usefdist}, part~\ref{u2}, we conclude 
that \mbox{$X \cap Y \leC Y \cap Z$}. 
Therefore, by Lemma~\ref{le:hi1}, part~\ref{eq} we have
\mbox{$\cC(X \cap Y) = \cC(X \cap Y \cap Z)$}.
But the same lemma implies \mbox{$\cC(X) = \cC(X \cap Y)$}.
Similarly, from \mbox{$X \leC Y$} and \mbox{$Z \leC Z$}, one concludes,
from Lemma~\ref{le:usefdist}, part~\ref{u2},
that \mbox{$X \cap Z \leC Y \cap Z$} and
\mbox{$\cC(X \cap Z) = \cC(X \cap Y \cap Z)$}.
Therefore \mbox{$\cC(X) = \cC(X \cap Z)$} and \mbox{$X \leC Z$}.
\QED
The following may help understand the meaning of the relation $\leC$, for
a distributive operation \cC.
\begin{theorem}
\label{the:help}
Let \cC\ be a distributive operation. The following three properties are 
equivalent.
\begin{enumerate} 
\item \label{p1}
\mbox{$X \leC Y$} 
\item \label{p2} there exists a set of formulas \mbox{$Y' \subseteq \Cn(Y)$}
such that \mbox{$X \subseteq  \cC(Y')$},
\item \label{p3}
there exists a set of formulas \mbox{$Y' \subseteq \Cn(Y)$}
such that \mbox{$\cC(X) =  \cC(Y')$}.
\end{enumerate}
\end{theorem} 
\proof     
Suppose \mbox{$X \leC Y$}. Then \mbox{$\Cn(X) \cap \Cn(Y)$} is a $Y'$ suitable
for property~\ref{p3}, by Lemma~\ref{le:hi1}.
It is clear that property~\ref{p3} implies property~\ref{p2}.
Suppose, now, that property~\ref{p2} holds.
By Lemma~\ref{le:hi1}, \mbox{$Y' \leC Y$}.
By the easy part of Theorem~\ref{the:distweak}, \mbox{$X \leC Y'$}.
We conclude, by Theorem~\ref{the:order}, that property~\ref{p1} holds.
\QED
We shall now fulfil our promise and show that distributive operations
are strongly cumulative. Notice we do not make any assumption on the 
language \cL.
The analogue result in the finitary setting appears in~\cite{KLMAI:89}.
In the infinitary setting, for classical propositional
calculus, it appears in~\cite{Mak:90}.
We prove a more general result, but the proof is similar.
\begin{theorem}
\label{the:diststr}
Any distributive operation is strongly cumulative.
\end{theorem}
\proof
Let \cC\ be a distributive operation.
Let \mbox{$X_{i} \subseteq \cC(X_{i+1})$}, for \mbox{$i = 0 , \ldots , n-1$},
where addition is understood modulo $n$.
We see that, since \cC\ is distributive, by the easy part of
Theorem~\ref{the:distweak} \mbox{$X_{i} \leC X_{i+1}$}, for any $i$.
Therefore, from Theorem~\ref{the:order} we conclude that 
\mbox{$X_{i+1} \leC X_{i}$}, for any $i$.
We conclude by Lemma~\ref{le:hi1}, part~\ref{anti}.
\QED
In~\cite{KLMAI:89}, a finitary strongly cumulative operation that is not
finitarily distributive (in the setting of classical propositional calculus)
was presented. Its smallest extension provides an example of a strongly 
cumulative operation that is not distributive.

\subsection{Extending a finitary distributive operation}
\label{subsec:extdist}
Now, we shall study the existence of distributive extensions for an
arbitrary distributive finitary operation.
As we have seen above, we must assume that the language \cL\ has 
disjunction for the notion of a distributive finitary operation
to make sense.
We shall therefore assume that \cL\ has disjunction.
By Lemma~\ref{le:basic}, part~\ref{bdis} (see the Appendix), then,  
the language \cL\ is admissible.
Our first result is negative.
\begin{theorem}
\label{the:neg}
The smallest extension of a distributive finitary operation is not, in general,
distributive.
\end{theorem}
\proof
Let, \cL\ be the classical propositional calculus on the infinite set
of variables: 
\mbox{$q , p_{1} , p_{2} , \ldots , p_{k} , \ldots$}.
Let \mbox{$\cF(A)$} be \mbox{$\Cn(A , q \ra p_{1})$}.           
Note that \mbox{$p_{1} \in \cF({q})$}. 
It is very easy to see that \cF\ is cumulative.
It is also easy, with the help of the disjunction that exists in \cL, to
show it is distributive.
Let \cC\ be the smallest extension of \cF. 
If $Y$ is the set of all $p_{i}$'s, we have 
\[
p_{1} \in \cC(Y) \cap \cC(q).
\]    
We shall show that, nevertheless, $p_{1}$ is not an element of 
\mbox{$\cC(\Cn(Y) \cap \Cn(q)$}. 
We claim, indeed, first that  \mbox{$\Cn(Y) \cap \Cn(q)$}
is not of the first type.        
For suppose $A$ is a finite set such that 
\begin{equation}
\label{eq:**}
A \subseteq  \Cn(\Cn(q) \cap \Cn(Y)) = \Cn(q) \cap \Cn(Y) \subseteq  \cF(A).
\end{equation}
Then, for any $i$, \mbox{$p_{i} \vee q \in \cF(A)$}, i.e., 
\mbox{$p_{i} \vee q \in \Cn(A , q \ra p_{1})$}. 
But it is not difficult to see that this implies that 
\mbox{$p_{i} \vee q \in \Cn(A)$} for any $i$.
Now, $q$ is not an element of $\Cn(A)$, since $q$ is not in $\Cn(Y)$, 
so there exists a world $m$ that satisfies $A$ and does not satisfy $q$. 
Since $A$ is finite, there is a $j$ such that $p_{j}$ does not appear 
in $A$. 
Let $m'$ be the world that differs from $m$ in at most $p_{j}$ and
in which $p_{j}$ is false.
Then $m'$ satisfies $A$ but satisfies neither $q$ nor $p_{j}$. 
A contradiction.
We have shown that there is no finite set $A$ such that (\ref{eq:**}) holds.
Therefore \mbox{$\Cn(Y) \cap \Cn(q)$} is not of the first type
and 
\[\begin{array}{lll}
\cC(\Cn(q) \cap \Cn(Y)) & = & \Cn(\Cn(q) \cap \Cn(Y)) \\
& = & \Cn(q) \cap \Cn(Y) .      
\end{array}
\]
But, $p_{1}$ is not an element of $\Cn(q)$.
\QED
Our next lemma is fundamental in our study of canonical extensions
of distributive finitary operations.
\begin{lemma}
\label{le:AB}
Suppose \cL\ has disjunction.
Let \cF\ be a distributive finitary operation.
If \mbox{$B \subseteqf \CF(X)$}, then there exists 
\mbox{$A \subseteqf \Cn(X)$} such that \mbox{$\cF(A) = \cF(B)$}.
\end{lemma}
\proof
By Lemma~\ref{le:tech}, there exists a finite subset $B'$ of $X$,
such that \mbox{$B \subseteqf \cF(B' \cup C)$} 
for any \mbox{$C \subseteqf \Cn(X)$}.
In particular \mbox{$B \subseteqf \cF(B')$}.
By the distributivity of \cF\ and the finitary version of
Lemma~\ref{le:usefdist}, part~\ref{u1}, \mbox{$\cF(B) = \cF(B \vee B')$}.
But \mbox{$B \vee B' \subseteqf \Cn(X)$} since $B'$ is a subset of $X$.
\QED
\begin{theorem}
\label{the:CFCFprime}
Suppose \cL\ has disjunction.
Let \cF\ be a distributive finitary operation.
The operation \CF\ is distributive and is the canonical extension of \cF.
\end{theorem}
\proof
To show that \CF\ is the canonical extension of \cF, we have to show 
that \CF\ is equal to its transform $\CF'$.
In view of Lemma~\ref{le:smaller}, we have to prove that,
for any $X$, \mbox{$\CF(X) \subseteq \CF'(X)$}.
Take an arbitrary element $a$ of the first set.
There is a finite subset $A$ of $X$, such that $a$ is an element
of $\cF(A , B)$ for any finite subset $B$ of $\Cn(X)$.
We must show that \mbox{$a \in \CF'(X)$}.
We shall take $A$ to be the $A$ of Definition~\ref{def:trans}
and show that \mbox{$a \in \CF(A , Y)$} for any \mbox{$Y \subseteq \CF(X)$}.
To show this, by Definition~\ref{def:CF}, it is enough to
show that \mbox{$a \in \cF(A , C)$} for any finite subset $C$
of $\Cn(A , Y)$.
Let $C$ be any such set.
Since \mbox{$\Cn(A , Y) \subseteq \CF(X)$}, $C$ is a finite subset of
$\CF(X)$, and so is \mbox{$A \cup C$}.
By Lemma~\ref{le:AB}, there is a finite subset $B$ of $\Cn(X)$ such that
\mbox{$\cF(B) = \cF(A , C)$}.
But, now \mbox{$A \subseteqf \cF(B)$} and, by cumulativity of \cF,
\mbox{$\cF(B) = \cF(A , B)$}.
But \mbox{$a \in \cF(A , B)$} and we conclude
\mbox{$a \in \cF(A , C)$}.

It is now time to show that \CF\ is cumulative. Notice that we do not assume
here that \cL\ has implication and cannot therefore
rely on Theorem~\ref{the:canext} to show that \CF\ is a cumulative extension.
By Lemmas~\ref{le:ext}, \ref{le:supra} and~\ref{le:cut}, we know \CF\ is a 
supraclassical
extension of \cF\ that satisfies Cut. Let us show it satisfies Cautious
Monotonicity.
Suppose \mbox{$Y \subseteq \CF(X)$}.
We must show \mbox{$\CF(X) \subseteq \CF(X , Y)$}.
Let \mbox{$a \in \CF(X)$}. By what we have just seen 
\mbox{$a \in \CF'(X)$} and therefore there exists a set \mbox{$A \subseteqf X$}
such that \mbox{$a \in \CF(A , Z)$} for any \mbox{$Z \subseteq \CF(X)$}.
But, since \mbox{$Y \subseteq \CF(X)$}, \mbox{$\Cn(X , Y) \subseteq \CF(X)$}.
We conclude that \mbox{$a \in \CF(A , B) = \cF(A , B)$} for any
\mbox{$B \subseteqf \Cn(X , Y)$}, and therefore
\mbox{$a \in \CF(X , Y)$}.

Let us now show that \CF\ is distributive.
By Lemma~\ref{le:basic}, part~\ref{bdis}, \cL\ is admissible and,
by Theorem~\ref{the:distweak}, part~\ref{w1} it is enough to show that
\mbox{$\CF(X) \cap \CF(Y) \subseteq$} \nolinebreak[3]
\mbox{$\CF(\Cn(X) \cap \Cn(Y))$}.
Suppose \mbox{$ a \in \CF(X) \cap \CF(Y)$}.
There is a set \mbox{$A \subseteqf X$} such that \mbox{$a \in \cF(A , B)$}
for any \mbox{$B \subseteqf \Cn(X)$}, and there is a set
\mbox{$A' \subseteqf Y$} such that \mbox{$a \in \cF(A' , C)$}
for any \mbox{$C \subseteqf \Cn(Y)$}.
Since \cL\ has disjunction, we may consider the set \mbox{$A \vee A'$}.
Clearly \mbox{$A \vee A' \subseteqf$} \nolinebreak[3]
\mbox{$\Cn(X) \cap \Cn(Y)$}.
To show that \mbox{$a \in \CF(\Cn(X) \cap \Cn(Y))$} we shall show
that \mbox{$a \in \cF(A \vee A' , B)$} for any
\mbox{$B \subseteqf$} \nolinebreak[3] \mbox{$\Cn(\Cn(X) \cap \Cn(Y)) =$}
\nolinebreak[3] \mbox{$\Cn(X) \cap \Cn(Y)$}.
But, by Distributivity,
\mbox{$\cF(A , B) \cap \cF(A' , B) \subseteq$} \nolinebreak[3]
\mbox{$\cF(A \vee A' , B)$}.
But, \mbox{$a \in \cF(A , B)$} and \mbox{$a \in \cF(A' , B)$}.
\QED

\subsection{Models for distributive operations}
\label{subsec:moddist}
We do not know of an interesting representation theorem for distributive
operations.
We shall summarize what we know.
It is easy to see that, 
any cumulative model the labeling function $l$ of which labels each state 
with a singleton (i.e., a single world) defines a distributive operation,
at least if the language \cL\ satisfies the following semantic property:
for any sets $X$, $Y$ of formulas, any world $m$ that satisfies 
all the formulas of \mbox{$\Cn(X) \cap \Cn(Y)$} satisfies either $X$ or $Y$.
A distributive operation that cannot be defined by such a model,
in the setting of classical propositional calculus,
is described by K.~Schlechta~\cite{Schle:91,Mak:90}.
It seems that there are two interesting open questions:
to find a model-theoretic characterization of distributive operations
and to find a proof-theoretic characterization of those 
operations that may be defined by cumulative (or ordered cumulative) models
in which each state is labeled by a single world.

\section{Deductive Operations}
\label{sec:ded}
\subsection{Introduction and Plan}
In this section we define a family of cumulative operations, deductive 
operations.
Deductive operations are essentially those that satisfy the infinitary
version of the S rule of~\cite{KLMAI:89}. They were shown to be the operations
representable by preferential structures in~\cite{FL:JELIA90}.
Deductivity is the property named {\em infinite conditionalization} 
in~\cite{FLM:89}, \cite{Mak:90} and~\cite{Schle:91}.
In~\ref{subsec:deddef}, we define both infinitary and finitary versions
of the Deductivity property and prove some first results about them.
In~\ref{subsec:dedcomp} we extensively compare the properties of Deductivity
and Distributivity. We show, that, under mild assumptions on \cL,
all deductive operations are distributive and that, under restrictive 
assumptions on the language \cL, deductive finitary operations coincide with 
distributive finitary operations.
In~\ref{subsec:dedext} we show, under mild hypotheses concerning
the language \cL, that the canonical extension of any deductive finitary 
operation \cF\ is equal to \CF\ and is a deductive operation. 
In Section~\ref{subsec:dedPoole}, we study one popular way of defining
inference operations, proposed by D.~Poole.
We show that, under weak assumptions on \cL, Poole systems define a strongly 
cumulative operation, finite Poole systems define an operation that coincide 
with the \CF\ of its restriction \cF\ to finite sets, 
Poole systems without constraints define a distributive operation and that
finite Poole systems without constraints
define a deductive operation that is the canonical extension of its
restriction to finite sets.
We provide an example of a distributive operation that is not deductive.
In~\ref{subsec:dedrep} we discuss models for deductive operations and
prove a representation theorem. 
This result is a non-trivial variation on, and a sharpening of the 
representation result
of~\cite{KLMAI:89} for preferential relations (Theorem 5.18).

\subsection{Definitions and important properties}
\label{subsec:deddef}
We shall now introduce the family of inference operations that
satisfy a property (called here Deductivity) that is the infinitary
analogue of the condition S of~\cite{KLMAI:89}.
Neither this property nor its finitary version need the presence of connectives
in the language \cL\ to be formulated.
Many results in the sequel, though, rely on the presence of connectives.
\begin{definition}
\label{def:deduct}
An inference operation \cC\ is {\em deductive} iff it is cumulative
and satisfies the following, for arbitrary \mbox{$X , Y \subseteq \cL$}:
\[
({\rm Deductivity})\ \ \cC(X , Y) \subseteq \Cn(X , \cC(Y)).
\]
A finitary inference operation \cF\ is {\em deductive} iff it is 
cumulative and satisfies the following, for arbitrary 
\mbox{$A , B \subseteqf \cL$}:
\[
({\rm Finitary \ Deductivity})\ \ \cF(A , B) \subseteq \Cn(A , \cF(B)).
\]
\end{definition}
Since, as is easily checked, any operation that satisfies Deductivity
satisfies Cut, we could have weakened the cumulativity requirement in
definition~\ref{def:deduct} to cautious monotonicity.
Notice that we do not require deductive operations to be distributive.
The property of Deductivity expresses the requirement that, if some formula
$a$ may be inferred from some assumptions $Y$ and some additional assumptions 
$X$, then, from $Y$ alone one could have inferred that if $X$ holds then
$a$ must hold. It looks very much like one half of the property defining
implication. Indeed, if \cL\ has implication and \cC\ is deductive, then
\mbox{$b \in \cC(X , a)$} implies \mbox{$a \ra b \in \cC(X)$}.
Deductivity is slightly stronger than this property, though, since it
encompasses also the case $a$ is not a formula but an (infinite) set
of formulas.  
The following provides a characterization of deductive operations.
\begin{theorem}
\label{the:dedchar}
Let \cC\ be a cumulative operation.
The following three properties are equivalent.
\begin{enumerate}
\item \label{pr3} The operation \cC\ is deductive,
\item \label{pr1} if \mbox{$Y \subseteq \cC(X)$}, then 
\mbox{$\cC(X) \subseteq \Cn(X , \cC(Y))$},
\item \label{pr2} if \mbox{$Y \leC X$}, then 
\mbox{$\cC(X) \subseteq \Cn(X , \cC(Y))$}.
\end{enumerate}
\end{theorem}
\proof
Suppose \cC\ is deductive. 
Let us show it satisfies condition~\ref{pr2}.
If \mbox{$Y \leC X$}, then, by Lemma~\ref{le:hi1}, part~\ref{eq},
we have \mbox{$\cC(\Cn(X) \cap \Cn(Y)) = \cC(Y)$}.
By Deductivity, though, 
\[\begin{array}{lll}
\cC(X) & = & \cC(X , \Cn(X) \cap \Cn(Y)) \\
& \subseteq & \Cn(X , \cC(\Cn(X) \cap \Cn(Y)).
\end{array}
\] 
Therefore \mbox{$\cC(X) \subseteq \Cn(X , \cC(Y))$}.
Let us now show that condition~\ref{pr2} implies condition~\ref{pr1}.
Suppose \mbox{$Y \subseteq \cC(X)$}.
Notice we cannot use here Theorem~\ref{the:distweak}, since \cC\ is not
assumed to be distributive.
But, by Cumulativity, we have \mbox{$\cC(X) = \cC(X , Y)$}.
By Lemma~\ref{le:hi1}, part~\ref{refl} \mbox{$Y \leC X \cup Y$} and, 
by property~\ref{pr2},
\mbox{$\cC(X , Y) \subseteq \Cn(X , Y , \cC(Y)) =$}
\mbox{$\Cn(X , \cC(Y))$}.
We conclude that property~\ref{pr1} is verified.
Let us show that property~\ref{pr1} implies Deductivity.
But, since \mbox{$Y \subseteq \cC(X , Y)$},
property~\ref{pr1} implies \mbox{$\cC(X , Y) \subseteq \Cn(X , Y , \cC(Y)) =$}
\mbox{$\Cn(X , \cC(Y))$}.
\QED
The following result is a key property of deductive and distributive 
operations. Notice no assumption on \cL\ is needed.
\begin{lemma}
\label{le:key}
Let \cC\ be a deductive and distributive operation.
If \mbox{$X \leC Y \leC Z$}, then \mbox{$Y \subseteq  \Cn(Z , \cC(X))$}.
\end{lemma}
\proof
Without loss of generality, one may assume that $X$, $Y$ and $Z$ are
theories.
Suppose \mbox{$X \leC Y \leC Z$}.
Since \cC\ is distributive we may apply Theorem~\ref{the:order}
to obtain \mbox{$X \leC Z$}.
By Lemma~\ref{le:usefdist}, part~\ref{u2}, using Distributivity again, from 
\mbox{$X \leC Y$}
and \mbox{$X \leC Z$}, we obtain
\mbox{$X \leC Y \cap Z$} and, by Lemma~\ref{le:hi1},
\mbox{$\cC(X) = \cC(X \cap Y \cap Z)$}.
But, by Deductivity,
\[
\cC(Y \cap Z) \subseteq
\Cn(Y \cap Z , \cC(X \cap Y \cap Z)) \subseteq 
\Cn(Z , \cC(X)).
\]
Since \mbox{$Y \subseteq \cC(Y \cap Z)$} we conclude that
\mbox{$Y \subseteq \Cn(Z , \cC(X))$}.
\QED
The following Theorem presents an easy result on monotonic deductive operations.
\begin{theorem}
\label{the:montrans}
Let \cC\ be a deductive operation such that, for any \mbox{$X \subseteq \cL$}, 
\mbox{$\cC(\emptyset) \subseteq \cC(X)$}.
Then \cC\ is monotonic and compact. Moreover, 
\mbox{$\cC(X) = \Cn(X , \cC(\emptyset))$}.
\end{theorem}
\proof
Let \cC\ be as above. By Deductivity,
\mbox{$\cC(X) \subseteq \Cn(X , \cC(\emptyset))$}.
By the other assumption, we have equality and Monotonicity is proven.
Compactness is easily verified.
\QED
Some more important properties of deductive operations will be described after
we clear up the relation between deductive and distributive operations.

\subsection{Deductive vs. Distributive operations} 
\label{subsec:dedcomp}
We shall now compare the two families of deductive and distributive
operations. 
Though the result of this comparison may well depend on the underlying \cL,
we shall see that, in typical cases, deductive operations form a strict 
sub-family of distributive operations, whereas deductive and distributive
finitary operations coincide. This is the case, for example, in the setting
of classical propositional calculus.
\begin{theorem}
\label{the:deddist}
If the language \cL\ is admissible, any deductive operation is distributive.
\end{theorem}
\proof
Let \cC\ be deductive.
By Theorem~\ref{the:wf}, it is enough to show that,
if \mbox{$Y \subseteq$} \nolinebreak[3] \mbox{$\cC(X)$}, then 
\mbox{$Y \leC X$}, for any theories $X , Y$.
By Theorem~\ref{the:dedchar}, \mbox{$\cC(X) \subseteq$} \nolinebreak[3] 
\mbox{$\Cn(X , \cC(X \cap Y)) \eqdef$} \nolinebreak[3] $Z$.
But \cL\ is admissible, and \mbox{$\Cn(X , Z) \cap \Cn(Y , Z) =$} 
\nolinebreak[3]
\mbox{$\Cn(X \cap Y , Z) =$} \nolinebreak[3] \mbox{$\Cn(Z) =$}
\nolinebreak[3] $Z$.
\QED
In Section~\ref{subsec:dedPoole}, we shall show that the converse does not
hold, when \cL\ is classical propositional calculus.
There is a result similar to Theorem~\ref{the:deddist} for finitary operations,
but, for the notion of a distributive finitary operation to make sense, we must
suppose the language \cL\ has disjunction.
Since the proof is similar to that 
of Theorem~\ref{the:deddist} it will be omitted.
\begin{theorem}
\label{the:findeddist}
If the language \cL\ has disjunction, 
any deductive finitary operation is distributive.
\end{theorem}
The next theorem provides a converse to Theorem~\ref{the:findeddist}
\begin{theorem}
\label{the:findistded}
If the language \cL\ has conjunction, disjunction and
classical negation,
any distributive finitary operation is deductive.
\end{theorem}
\proof
The proof is the abstract and infinitary version of the proof of Lemma 5.2.
of~\cite{KLMAI:89}.
The assumption that \cL\ has disjunction is needed only
for the notion of a distributive finitary operation to make sense.
Suppose \cL\ is as assumed and \cF\ is distributive.
We must show that, for any \mbox{$A , B \subseteqf \cL$},
\mbox{$\cF(A , B) \subseteq \Cn(A , \cF(B))$}.
It is clear that
\mbox{$\cF(A , B) \subseteq \Cn(A , \cF(A , B))$}.
Since, by parts~\ref{bclassneg} and~\ref{bintneg} of Lemma~\ref{le:basic},
\mbox{$\Cn(A , \neg \chi_{A}) = \cL$}, we also have:
\mbox{$\cF(A , B) \subseteq$} \mbox{$\Cn(A , \cF(\neg \chi_{A} , B))$}.
Therefore, 
\[
\cF(A , B) \subseteq \Cn(A , \cF(A , B)) \cap \Cn(A , 
\cF(\neg \chi_{A} , B)).
\]
By part~\ref{bdis} of Lemma~\ref{le:basic},
\bigskip

\noindent $\Cn(A , \cF(A , B)) \cap \Cn(A , \cF(\neg \chi_{A} , B)) = $

\hfill{$\Cn(A , \cF(A , B) \cap \cF(\neg \chi_{A} , B)).$}
\bigskip

\noindent Since \cF\ is distributive,
\bigskip

\noindent $\cF(A,B)\!\!\cap\!\!\cF(\neg\!\!\chi_{A},B)\!\!\subseteq\!\!\cF(\Cn(A,B)\!\!\cap\!\!\Cn(\neg\chi_{A},B))\!\!=$

\hfill{$\cF(B , \Cn(A) \cap \Cn(\neg \chi_{A})).$}
\bigskip

\noindent But, since \cL\ has classical negation, we may apply part~\ref{bnewclassneg}
of Lemma~\ref{le:basic} to conclude that
$\Cn(A) \cap \Cn(\neg \chi_{A}) = \Cn(\emptyset)$
and $\cF(B , \Cn(A) \cap \Cn(\neg \chi_{A})) = \cF(B)$.
We conclude that
$\cF(A , B) \subseteq \Cn(A , \cF(B))$.
\QED
We end up this section with a partial converse to 
Theorems~\ref{the:deddist} and \ref{the:montrans}.
\begin{theorem}
\label{the:distded}
If the language \cL\ has conjunction, disjunction and
classical negation, any distributive inference operation that is both monotonic
and compact is deductive and a translation of \Cn.
\end{theorem}
\proof
Let \cC\ be a distributive, monotonic and compact operation.
Suppose \mbox{$a \in \cC(X , Y)$}.
Since \cC\ is compact, there are \mbox{$A \subseteqf X$}, 
\mbox{$B \subseteqf Y$} such that \mbox{$a \in \cC(A , B)$}.
By Theorem~\ref{the:findistded}, \cC\ is finitarily deductive
and \mbox{$\cC(A , B) \subseteq \Cn(A , \cC(B))$}.
By Monotonicity of \Cn\ and \cC\ we see that
\mbox{$\Cn(A , \cC(B)) \subseteq \Cn(X , \cC(Y))$}.
\QED

\subsection{Canonical extensions of deductive operations}
\label{subsec:dedext}
We want here to deal with the question of whether all deductive finitary
operations have deductive extensions.
We do not know whether this is the case if one assumes only that \cL\ has 
disjunction.
\begin{theorem}
\label{the:CFcanext}
Assume \cC\ has implication and disjunction.
Let \cF\ be a deductive finitary operation.
The canonical extension of \cF\ is deductive and equal to \CF.
\end{theorem}
\proof
By Theorem~\ref{the:findeddist}, the operation \cF\ is distributive.
Theorem~\ref{the:CFCFprime} therefore asserts that \CF\ is indeed a
distributive extension of \cF\ that is its canonical extension.
We are left to show that \CF\ satisfies Deductivity.
Suppose \mbox{$a \in \CF(X , Y)$}.
We shall show that \mbox{$a \in \Cn(X , \CF(Y))$}.
We know that there is \mbox{$A \subseteqf X \cup Y$} such that
\mbox{$a \in \cF(A , B)$} for any \mbox{$B \subseteqf \Cn(X , Y)$}.
Since \cF\ is deductive, 
$a \in \Cn(A , \cF(B))$ for any $B \subseteqf \Cn(X , Y)$.
But \cL\ has implication and we may use part~\ref{bconimp3}
of Lemma~\ref{le:basic} to conclude that
\[
a \in \Cn(A , \bigcap_{B \subseteqf \Cn(X , Y)} \cF(B)).
\]
It is left to us to show that
$$\bigcap_{B \subseteqf \Cn(X , Y)} \cF(B) \subseteq \CF(Y).$$
But, if $b$ is an element of the first set, one may take \mbox{$A = \emptyset$}
to show that $b$ is in the second set.
\QED
In~\cite{Freund:92}, a slightly stronger result is proven, in the case \cL\
a propositional calculus: the transform of any distributive operation
is deductive.

\subsection{Poole Systems}
\label{subsec:dedPoole}
In~\cite{Poole:88}, D.~Poole defined a formalism for nonmonotonic reasoning.
This formalism may be considered as a method for defining inference operations.
We shall quickly recall here the Definitions of~\cite{Poole:88}.
Then, we shall show that Poole systems with constraints define strongly 
cumulative operations,
that finite Poole systems with constraints define an operation \cC\ that 
is the canonical extension of its restriction to finite sets,
that Poole systems without constraints define distributive operations
and that finite Poole systems without constraints define deductive operations. 
We shall then provide an example of a distributive operation that is not 
deductive.

Previous works, and in particular~\cite{Mak:90} and~\cite{FLM:89}, 
have assumed that the language \cL\ is classical propositional calculus.
We shall not make this assumption.
All the results presented here are, therefore, technically new.
Most of the proofs, though, owe a lot to~\cite{Mak:90} and~\cite{FLM:89}.
It is difficult to sort out exactly the credits, but D.~Makinson played,
with the authors, a major role in the elaboration of those results.
\begin{definition}
\label{def:Poole}
A Poole system (with constraints) is a pair 
\mbox{$\langle D , K \rangle$} of sets of formulas. 
The set $D$ is called the set of defaults. 
The set $K$ is the set of constraints.
The system \mbox{$\langle D , K \rangle$} is said to be {\em finite}
iff the set of defaults $D$ is finite (the set of constraints may be infinite).
It is said to be without constraints iff the set of constraints $K$ is empty.
\end{definition}
Suppose such a Poole system \mbox{$\langle D , K \rangle$} is given.
Consider a set $X$ of formulas.
A subset \mbox{$A \subseteq D$}
is said to be a {\em basis} for $X$ iff \mbox{$\Cn(X , A , K) \neq \cL$}
and $A$ is a maximal subset of $D$ with this property.
We shall denote the set of all bases for $X$ by \mbox{$\cB(X)$}.
The inference operation defined by \mbox{$\langle D , K \rangle$} is now:
\begin{equation}
\label{eq:Poole}
\cC(X) \eqdef \bigcap_{A \in \cB(X)} \Cn(X , A).
\end{equation}
\begin{lemma}
\label{le:ante}
If \cC\ is defined as in Equation~(\ref{eq:Poole}), 
\begin{enumerate}
\item \label{incl} the operation \cC\ is supraclassical
and 
\item \label{ante} for any $B \in \cB(X)$,
$\Cn(\cC(X) , B) = \Cn(X , B)$.
\end{enumerate}
\end{lemma}
\proof
Property~\ref{incl} is obvious from Equation~(\ref{eq:Poole}).
For property~\ref{ante}, since 
\mbox{$\cC(X) \subseteq$} \nolinebreak[3] \mbox{$\Cn(X , B)$},
\mbox{$\Cn(\cC(X) , B) \subseteq$} \nolinebreak[3]
\mbox{$\Cn(\Cn(X , B) , B) =$} \nolinebreak[3]
\mbox{$\Cn(X , B)$}.
The inclusion in the other direction follows from property~\ref{incl}.
\QED
\begin{lemma}
\label{le:prelimPoole}
Assume the language \cL\ has contradiction.
If \mbox{$X \subseteq Y$} and \mbox{$B \in \cB(Y)$}, then, there is a 
\mbox{$B' \in \cB(X)$} such that \mbox{$B \subseteq B'$}.
\end{lemma}
\proof
Suppose \mbox{$X \subseteq Y$} and \mbox{$B \in \cB(Y)$}. \linebreak[3]
Clearly \linebreak[2] \mbox{$\Cn(X , B , K) \subseteq$} \nolinebreak[3]
\mbox{$\Cn(Y , B , K) \neq$} \nolinebreak[3] \cL. \linebreak[3]
Since the language \cL\ has contradiction, \linebreak[2] one may show, 
\linebreak[2] using 
part~\ref{bcontr} of Lemma~\ref{le:basic} that, \linebreak[2] if we have a
chain $B_{\alpha}$ for ordinals \mbox{$\alpha < \gamma$}, such that, for every 
\mbox{$\alpha < \beta < \gamma$}, \linebreak[2]
\mbox{$B_{\alpha} \subseteq B_{\beta}$}, and
\mbox{$\Cn(X , B_{\beta} , K) \neq \cL$}, \linebreak[2] then  
\mbox{$\Cn(X , \bigcup_{\beta < \gamma} B_{\beta} , K) \neq \cL$}.
One may, therefore build such a chain starting from $B$, until one obtains
a basis for $X$.
\QED
\begin{lemma}
\label{le:baXCX}
Assume the language \cL\ has contradiction.
If \cC\ is defined as in Equation~(\ref{eq:Poole}), then
\mbox{$\cB(X) = \cB(\cC(X))$}.
\end{lemma}
\proof
It follows easily from Lemma~\ref{le:ante} that any basis $B$ for
$X$ is a basis for $\cC(X)$.
Indeed, in view of part~\ref{incl} of Lemma~\ref{le:ante}, it is enough to
show that \mbox{$\Cn(\cC(X) , B , K) \neq \cL$}, which follows from
part~\ref{ante}.
Let, now, $B$ be a basis for $\cC(X)$.
By Lemma~\ref{le:baXCX}, there is a basis $B'$ for $X$ such that
\mbox{$B \subseteq B'$}.
By Lemma~\ref{le:ante}, \mbox{$\Cn(\cC(X) , B' , K) =$} \nolinebreak[3]
\mbox{$\Cn(X , B' , K) \neq$} \nolinebreak[3] \cL.
By the maximality of $B$, \mbox{$B' = B$} and we conclude that any basis
for $\cC(X)$ is a basis for $X$.
\QED
\begin{theorem}[Makinson]
\label{the:Mak1}
Assume the language \cL\ has contradiction.
The inference operation defined by any Poole system is strongly cumulative.
\end{theorem}
\proof
Notice that, in~\cite{Mak:90}, the same result is proved when
the setting is classical propositional calculus.
We assume much less on \cL.
We know from Lemma~\ref{le:ante} that \cC\ is supraclassical.
Suppose \mbox{$X_{i} \subseteq \cC(X_{i+1})$} for 
\mbox{$i = 0 , \ldots , n-1$},
where addition is understood modulo $n$.
First, we claim that, for any $i , j$, \mbox{$\cB(X_{i}) = \cB(X_{j})$}.
Suppose, indeed, $B_{i}$ is a basis for $X_{i}$.
By Lemma~\ref{le:baXCX}, $B_{i}$ is a basis for $\cC(X_{i})$.
Since, \mbox{$X_{i-1} \subseteq \cC(X_{i})$}, by Lemma~\ref{le:prelimPoole},
there is a basis $B_{i-1}$ for $X_{i-1}$ such that 
\mbox{$B_{i} \subseteq B_{i-1}$}. Since addition is modulo $n$,
there is a basis $B'$ for $X_{i}$ such that \mbox{$B_{i} \subseteq B'$}.
But $B_{i}$ is a basis for $X_{i}$, and we conclude that,
for any $i , j$, \mbox{$B_{i} = B_{j}$}, therefore $B_{i}$ is a basis
for $X_{j}$. Let \cB\ be the set of all bases for $X_{i}$ (or $X_{j}$).
By Lemma~\ref{le:ante}, for any $i$, for any $B$ in $\cB$,
\mbox{$\Cn(\cC(X_{i}) , B) =$} \nolinebreak \mbox{$\Cn(X_{i} , B)$}.
Since \mbox{$X_{i} \subseteq$} \nolinebreak[3] \mbox{$\cC(X_{i+1})$}, we have
\[\begin{array}{lll}
\cC(X_{i}) & = & \bigcap_{B \in \cB} \Cn(X_{i} , B)  \\
&\subseteq &\bigcap_{B \in \cB} \Cn(\cC(X_{i+1}) , B) \\
& = & \bigcap_{B \in \cB} \Cn(X_{i+1} , B) = \cC(X_{i+1}).
\end{array}
\]
We conclude easily that \mbox{$\cC(X_{i}) = \cC(X_{j})$}.
\QED
The following result suggests that our definition of the canonical extension
of a finitary operation is indeed natural.
\begin{theorem}
\label{the:cCcF}
Assume the language \cL\ has contradiction.
Let \cC\ be the inference operation defined by a {\em finite} Poole system. 
Let \cF\ be the restriction of \cC\ to finite sets.
The operation \cC\ is the canonical extension of \cF, and equal to \CF.
\end{theorem}
\proof
In view of Lemma~\ref{le:smaller}, it is enough for us to show that,
for any \mbox{$X \subseteq \cL$}, we have
\mbox{$\CF(X) \subseteq \cC(X) \subseteq$} \mbox{$\CF'(X)$}.
Let us first show that \mbox{$\CF(X) \subseteq \cC(X)$}.
We must show that, if
\mbox{$a \not \in \cC(X)$}, there exists no \mbox{$A \subseteqf X$}
such\ that \mbox{$a \in \cC(A , C)$}, for any \mbox{$C \subseteqf \Cn(X)$}.
Suppose $a$ is not an element of $\cC(X)$. 
Then there exists a basis $B$ for $X$, such that $a$ is not in $\Cn(X , B)$.  
For any \mbox{$d \in D - B$}, we have \mbox{$\Cn(X , B , d , K) = \cL\ $}, 
because of the maximality of $B$. 
By part~\ref{bcontr} of Lemma~\ref{le:basic}, there exists a 
finite subset $X'$ of $X$ such that 
\mbox{$\Cn(X' , B , d , K) = \cL$}. 
Let $Y$ be the union of all the $X'$ obtained for each $d$.
Since $D$ is finite, the set $Y$ is finite.
It is clear that $B$ is a basis for $Y$, and therefore also a basis for 
\mbox{$Y \cup A$}
for any subset $A$ of $X$.
We want now to show that there is no \mbox{$A \subseteqf X$} such that
\mbox{$a \in \cC(A , C)$} for any \mbox{$C \subseteqf \Cn(X)$}.
Let $A$ be an arbitrary finite subset of $X$.
It is enough to show that \mbox{$a \not \in \cC(A , Y)$}.
But $B$ is a basis for \mbox{$A \cup Y$}, and 
\mbox{$a \not \in \Cn(A , Y , B)$}
since 
\mbox{$a \not \in \Cn(X , B)$}.
We conclude that $a$ is not in \mbox{$\cC(A , Y)$}.

Let us, now, show that \mbox{$\cC(X) \subseteq \CF'(X)$}.
Suppose \mbox{$a \in \cC(X)$}.
For any basis $B$ for $X$, \mbox{$a \in \Cn(X , B)$}.
By compactness of \Cn, for any basis $B$ for $X$, there is a finite
subset $A_{B}$ such that \mbox{$a \in \Cn(A_{B} , B)$}.
Let $A_{0}$ be the union of all those $A_{B}$.
Since the set of defaults is finite, $A_{0}$ is a finite subset of $X$, and,
for any basis $B$ for $X$, \mbox{$a \in \Cn(A_{0} , B)$}.
Let, now, \cE\ be the set of all subsets $E$ of $D$ 
(the finite set of defaults)
such that \mbox{$\Cn(X , E , K) = \cL$}.
By part~\ref{bcontr} of Lemma~\ref{le:basic}, 
for any \mbox{$E \in \cE$} there is a finite
subset $A_{E}$ of $X$ such that \mbox{$\Cn(A , E , K) = \cL$}.
Let $A_{1}$ be the union of all those $A_{E}$ for \mbox{$E \in \cE$}.
The set $A_{1}$ is a finite union of finite sets and is therefore a finite
subset of $X$, and, for any \mbox{$E \in \cE$}, 
\mbox{$\Cn(A_{1} , E , K) = \cL$}.

Let \mbox{$A \eqdef A_{0} \cup A_{1}$}.
The set $A$ is a finite subset of $X$.
We must show that, for any \mbox{$C \subseteq \cC(X)$},
one has \mbox{$a \in \cC(A , C)$}.
Let $C$ be an arbitrary subset of $\cC(X)$ and $B$ an arbitrary basis for 
$A \cup C$. It is enough to show that \mbox{$a \in \Cn(A , C , B)$}.
But \mbox{$\Cn(A_{1} , B , K) \neq \cL$} and therefore
\mbox{$B \not \in \cE$} and \mbox{$\Cn(X , B , K) \neq \cL$}.
There is, therefore, some
basis $B'$ for $X$ such that \mbox{$B \subseteq B'$}.
Since \mbox{$C \subseteq \cC(X)$}, we have
\mbox{$C \subseteq \Cn(X , B')$} and therefore 
\mbox{$\Cn(X , C , B' , K) = \Cn(X , B' , K) \neq \cL$} and
\mbox{$\Cn(A , C , B' , K) \neq \cL$}. We conclude that \mbox{$B' = B$}
and $B$ is a basis for $X$.
Therefore \mbox{$a \in \Cn(A_{0} , B$} and \mbox{$a \in \Cn(A , C , B)$}.
\QED
The result above should be compared with Theorem~\ref{the:CFCFprime}.
Here, the operation \cF\ is not always distributive, as has been shown
in~\cite{Mak:90}.
Poole systems without constraints, i.e., \mbox{$K = \emptyset$},
however, typically, define distributive operations.
\begin{theorem}[Makinson]
\label{the:Mak2}
Assume the language \cL\ has contradiction and is admissible.
The inference operation defined by any Poole system without constraints is 
distributive.
\end{theorem}
\proof
By Theorem~\ref{the:wf}, it is enough to show that the operation \cC\
defined by such a system is weakly distributive.
Let us examine, first, the bases for \mbox{$\Cn(X) \cap \Cn(Y)$}.
Let $B$ be such a basis. If $B$ is consistent with $X$ it is a basis
for $X$ and if it is consistent with $Y$ it is a basis for $Y$.
Since \cL\ is admissible, \mbox{$\Cn(X , B) \cap \Cn(Y , B) =$}
\nolinebreak[3] \mbox{$\Cn( \Cn(X) \cap \Cn(Y) , B)$} and, therefore,
$B$ cannot be  inconsistent with both $X$ and $Y$.
Therefore, a basis $B$ for \mbox{$\Cn(X) \cap \Cn(Y)$}
is a basis for both $X$ and $Y$, or a basis for $X$ such that
\mbox{$\Cn(Y , B) = \cL$} or a basis for $Y$ such that
\mbox{$\Cn(X , B) = \cL$}.
One concludes, since \cL\ is admissible, that
\begin{eqnarray*}
\cC(X) \cap \cC(Y) = \!\! \bigcap_{B \in \cB(X)} \!\!\! \Cn(X , B)
\cap \bigcap_{B \in \cB(Y)}\!\!\! \Cn(Y , B) \\ \subseteq
\bigcap_{B \in \cB(\Cn(X) \cap \Cn(Y))}\!\!\!\!\! \Cn(\Cn(X) \cap \Cn(Y) , B).
\end{eqnarray*}
\QED
Our next result shows that finite Poole systems without constraints
define deductive operations.
\begin{theorem}
\label{the:Pooded}
Assume the language \cL\ has contradiction and is admissible.
Let \cC\ be the inference operation defined by a {\em finite} Poole system 
without constraints.
The operation \cC\ is deductive.
\end{theorem}
\proof
We know from Theorem~\ref{the:Mak1} that \cC\ is cumulative.
We must show that it satisfies Deductivity.
We must show that \mbox{$\cC(X , Y) \subseteq \Cn(X , \cC(Y))$}.
Notice that, since there are only finitely many defaults, there are
only finitely many bases.
Since \cL\ is admissible, we have
\begin{eqnarray*}
\Cn(X , \cC(Y)) & = & \Cn(X , \bigcap_{B \in \cB(Y)} \Cn(Y , B)) \\
& = & \bigcap_{B \in \cB(Y)} \Cn(X , \Cn(Y , B)) \\
& = & \bigcap_{B \in \cB(Y)} \Cn(X , Y , B).
\end{eqnarray*}
Suppose \mbox{$a \in \cC(X , Y)$}. We shall show that, for any base $B$
for $Y$, \mbox{$a \in \Cn(X , Y , B)$}. Let $B$ be an arbitrary basis
for $Y$. Either \mbox{$\Cn(Y , X , B) = \cL$} or $B$ is a basis for 
\mbox{$Y \cup X$}.
In any case, \mbox{$a \in \Cn(X , Y , B)$}.
\QED
As we prove now, Theorem~\ref{the:Mak2} cannot be improved upon, in
the sense that, even in the setting of classical propositional calculus,
there is a Poole system without constraints that defines an operation
that is not deductive.
This provides an operation that is distributive but not deductive.
A number of such operations have been proposed. D.~Makinson seems to have
provided the first one. The most interesting is probably the one
provided by K.~Schlechta and mentioned in Section~\ref{subsec:moddist};
Theorem~\ref{the:fullcomp} shows indeed that it is not deductive.
Independently of these proposals, A.~Brodsky and R.~Brofman offered
the following.

Let \cC\ be classical propositional calculus and let $D$ be the following 
infinite set: \mbox{$p_{0}$}, \mbox{$p_{1} \wedge \neg p_{0}$},
\mbox{$p_{2} \wedge \neg p_{1} \wedge \neg p_{0}$}, 
\mbox{$\ldots , p_{i} \wedge \neg p_{i-1} \wedge \ldots \wedge \neg p_{0} , 
\ldots $}.
Clearly any two different elements of $D$ are inconsistent and therefore
a basis may have at most one element.
Now let \cC\ be the (distributive) operation defined by the Poole system
\mbox{$\langle D , \emptyset \rangle$}.
Let $X$ be the infinite set: \mbox{$p_{0} \ra q$}, \mbox{$p_{1} \ra q$},
\mbox{$\ldots , p_{i} \ra q , \ldots$}.
Clearly the bases for $X$ are exactly all the singletons of $D$ and
therefore \mbox{$q \in \cC(X)$}.
To show that \cC\ is not deductive we shall show that
\mbox{$q \not \in \Cn(X , \cC(\emptyset))$}.
Indeed we claim that $\cC(\emptyset)$ is the set of all tautologies, 
$\Cn(\emptyset)$.
Clearly the bases for $\emptyset$ are exactly all the singletons of $D$.
Suppose \mbox{$a \in \Cn(d)$} for every default \mbox{$d \in D$}.
Then $a$ may be false only in propositional models in which all the
$p_{i}$'s are false. Since $a$ refers to only a finite number of variables,
$a$ must be a tautology.
Therefore \mbox{$\Cn(X , \cC(\emptyset)) = \Cn(X)$}.
But \mbox{$q \not \in \Cn(X)$}.  

\subsection{Full Models and Representation Theorem}
\label{subsec:dedrep}
In this section we shall provide a representation theorem, showing that
the family of deductive operations is exactly the family of all operations
that are defined by certain models.
The reader remembers that in Section~\ref{subsec:moddist}, we noticed that
cumulative models that label states with singletons define distributive
operations. But this family was found to be too small to define all
distributive operations. Since, under weak assumptions on \cL, all
deductive operations are distributive, it is only natural we ask whether
cumulative models that label states with singletons may define all 
deductive operations. We shall provide a positive answer to that question.
Unfortunately, the operation defined by such a model is not always deductive.
We shall define a special class of such models that is large enough
to be able to define all deductive operations but small enough that
all operations it defines are deductive.
\begin{definition}
\label{def:full}
A cumulative ordered model \mbox{$W = \langle S , l , \prec \rangle$}
is said to be a full model iff
\begin{enumerate}
\item for every \mbox{$s \in S , l(s)$} contains a single world 
(from now one we shall identify $l(s)$ with its unique element) and
\item (fullness property) for any set \mbox{$X \subseteq \cL$} and any world 
\mbox{$m \in \cU$}
that satisfies all the formulas of $\CW(X)$, but does not satisfy all the
formulas of \cL, there is a state $s$,
minimal in $\widehat{X}$, such that \mbox{$l(s) = m$}.
\end{enumerate}
\end{definition}
Notice that we require the relation $\prec$ to be a strict partial order,
though, even without this assumption, the operation defined by a model
is deductive.
The second condition is difficult to check on
specific models.
The soundness result that follows makes no assumption on \cL.
\begin{theorem}
\label{the:soundfull}
If \mbox{$W = \langle S , l , \prec \rangle$} is a full model,
the operation \CW\ is deductive.
\end{theorem}
\proof
The reader will notice we do not use the fact that $\prec$ is a partial order.
We know from Theorem~\ref{the:cumsou}, that \CW\ is cumulative.
It is left to us to show that it satisfies Deductivity.
Let \mbox{$X , Y \subseteq \cL$}.
Suppose \mbox{$a \not \in \Cn(X , \cC(Y))$}.
Then, there is a world $m$ that satisfies $X$ and $\cC(Y)$ but
does not satisfy $a$.
Since $m$ satisfies $\cC(Y)$, by the fullness property, there is a state
$s$ minimal in $\widehat{Y}$ such that \mbox{$l(s) = m$}.
But $s$ is minimal in $\widehat{Y}$ and satisfies $X$.
Therefore it is minimal in $\widehat{X \cup Y}$.
This implies that \mbox{$a \not \in \cC(X , Y)$}.
\QED
The following is an easy corollary, in view of Theorem~\ref{the:deddist}.
Notice that no restrictive semantic assumption, as was formulated in
our discussion of Section~\ref{subsec:moddist}, is needed here.
\begin{corollary}
\label{co:fulldist}
Assume the language \cL\ is admissible.
The operation defined by a full model is distributive.
\end{corollary}
We shall now proceed to the proof of the representation theorem.
We shall assume that \cL\ is admissible.
From now on, \cC\ will be a fixed deductive operation on \cL.
We know from Theorem~\ref{the:deddist} that \cC\ is distributive.
We shall build a full model that defines \cC.
The construction is very similar to the one appearing 
in~\cite[Section 5.3]{KLMAI:89}.
We shall say that a world $m$ is normal for a set \mbox{$X \subseteq \cL$}
iff $m$ satisfies all the formulas of $\cC(X)$.
We proceed now to the construction of a full model 
\mbox{$W = \langle S , l , \prec \rangle$} that defines \cC. 
The set $S$ is taken to be the set of all pairs \mbox{$(m , X)$}
where $X$ is a set of formulas and $m$ is a normal world for $X$. 
The labelling function $l$ is the projection on the first coordinate and the 
strict partial order $\prec$ is defined by: 
\mbox{$(m , X) \prec (n , Y)$} iff 
\mbox{$X \leC Y$} and $m$ does not satisfy all the formulas of $Y$. 
One may notice immediately that this ensures that any state of the
form \mbox{$(m , X)$} is minimal in $\widehat{X}$.   
\begin{lemma}
\label{le:strict}
The relation $\prec$ is a strict partial order.
\end{lemma}
\proof
The relation is clearly irreflexive, so all we have to check is that 
it is transitive. Suppose hence that \mbox{$(m , X) \prec (n , Y)$} and 
\mbox{$(n , Y) \prec (p , Z)$}. 
Theorem~\ref{the:order} implies that \mbox{$X \leC Z$}. 
We have to show that $m$ does not satisfy $Z$.
But, by Lemma~\ref{le:key}, \mbox{$Y \subseteq \Cn(Z , \cC(X))$}.
The world $m$ does not satisfy $Y$ but satisfies $\cC(X)$, therefore it
does not satisfy $Z$.
\QED
\begin{lemma}
\label{le:charmin}
Let $X$, $Y$ be sets of formulas and $m$ a normal world for $X$.
The three propositions that follow are equivalent.
\begin{enumerate}
\item \label{minim}
The state \mbox{$(m , X)$} is minimal in $\widehat{Y}$.
\item \label{str}
The world $m$ satisfies $Y$ and $X$ is not a subset of
\mbox{$\Cn(Y , \cC(\Cn(X) \cap \Cn(Y)))$}.
\item \label{leC}
The world $m$ satisfies $Y$ and \mbox{$X \leC Y$}.
\end{enumerate}
\end{lemma}
\proof
Let $Z$ stand for \mbox{$\cC(\Cn(X) \cap \Cn(Y))$}.
Let us show that property~\ref{minim} implies property~\ref{str}.
Suppose \mbox{$(m , X)$} is minimal in $\widehat{Y}$.
Clearly $m$ satisfies $Y$.
If $X$ was a subset of
\mbox{$Z \eqdef \Cn(Y , Z)$},
there would be a world $n$, that satisfies $Z$, and is therefore
normal for \mbox{$\Cn(X) \cap \Cn(Y)$}, but does not satisfy $X$.
The pair \mbox{$(n , \Cn(X) \cap \Cn(Y))$} would therefore be a state of $S$
and, by Lemma~\ref{le:hi1}, part~\ref{refl}, we would have 
\mbox{$(n , \Cn(X) \cap \Cn(Y)) \prec (m , X)$},
a contradiction with the minimality of \mbox{$(m , X)$}.
Let us show now that property~\ref{str} implies property~\ref{leC}.
If \mbox{$X \subseteq \Cn(Y , Z$},
since \mbox{$\Cn(X) \cap \Cn(Y) \subseteq Z$}, we conclude by
Lemma~\ref{le:nadm} that \mbox{$X \subseteq Z$}, i.e. \mbox{$X \le Y$}.
Let us show now that property~\ref{leC} implies property~\ref{minim}.
Suppose $m$ satisfies $Y$ and \mbox{$X \leC Y$}.
The state \mbox{$(m , X)$} is in $\widehat{Y}$.
Suppose \mbox{$(n , W) \prec (m , X)$}.
We have, by Lemma~\ref{le:key}, \mbox{$X \subseteq \Cn(Y , \cC(W))$}.
Since $n$ is a normal world for $W$ that does not satisfy $X$, it does not
satisfy $Y$.
\QED
\begin{corollary}
\label{co:smooth}
The model $W$ satisfies the smoothness condition. It is therefore a
cumulative model.
\end{corollary}
\proof
By Lemma~\ref{le:charmin}, if $(m , X)$ is an element of $\widehat{Y}$
and is not minimal in this set, then 
\mbox{$X \not \subseteq$} \mbox{$\Cn(Y , \cC(\Cn(X) \cap \Cn(Y)))$}.
Therefore, there is a world $n$ that is normal for \mbox{$\Cn(X) \cap \Cn(Y)$},
satisfies $Y$ but does not satisfy $X$.
The state \mbox{$(n , \Cn(X) \cap \Cn(Y))$} is therefore, 
by Lemma~\ref{le:charmin} a minimal state of $\widehat{Y}$.
But clearly, \mbox{$(n , \Cn(X) \cap \Cn(Y)) \prec (m , X)$}.
\QED
\begin{lemma}
\label{le:CWC}
The operation \CW is equal to \cC.
\end{lemma}
\proof
Let $X$ be a set of formulas.
Suppose that \mbox{$\CW(X) \not \subseteq \cC(X)$}.
There would be a world $m$, that is normal for $X$ and does not 
satisfy $\CW(X)$.
But the pair $(m , X)$ would be a minimal state of $\widehat{X}$
by Lemma~\ref{le:charmin} and, by definition of \CW\ the world $m$
satisfies all formulas of $\CW(X)$. A contradiction.

Suppose now that \mbox{$\cC(X) \not \subseteq \CW(X)$}.
There is therefore a state $(m , Y)$ in $W$, minimal in $\widehat{X}$,
such that $m$ does not satisfy all formulas of $\cC(X)$.
But, by Lemma~\ref{le:charmin}, \mbox{$Y \le X$}.
By Theorem~\ref{the:dedchar}, then, 
\mbox{$\cC(X) \subseteq \Cn(X , \cC(Y))$}.
But $m$ satisfies $X$ and $\cC(Y)$, but not $\cC(X)$.
A contradiction.
\QED
\begin{corollary}
\label{co:full}
The model $W$ is a full model.
\end{corollary}
\proof
Suppose $m$ satisfies $\CW(X)$. By Lemma~\ref{le:CWC} it satisfies
$\cC(X)$ and $(m , X)$ is a state of $W$, and clearly, from the definition 
of $\prec$ or Lemma~\ref{le:charmin}, \mbox{$(m , X)$} is minimal in 
$\widehat{X}$.
\QED
We may now conclude. 
\begin{theorem}
\label{the:fullcomp}
Assume \cL\ is admissible.
Any deductive inference operation is defined by some full model.
\end{theorem}
\begin{theorem}
\label{the:fullrep}
Assume \cL\ is admissible.
An inference operation is deductive iff it is defined by some full model.
\end{theorem}
\proof
By Theorems~\ref{the:soundfull}, \ref{the:deddist} and~\ref{the:fullcomp}.
\QED

\section{Rational Inference Operations}
\label{sec:rat}
\subsection{Introduction and Plan}
\label{subsec:ratintro}
The families of operations described in the previous sections
provide a rich formal setting in which one may study nonmonotonic
reasoning. But, even if one is of the opinion that reasonable
nonmonotonic systems should implement a deductive operation, one may
ask whether any deductive operation provides a bona fide reasonable
nonmonotonic reasoning system.
In~\cite{KLMAI:89} and in~\cite{LMAI:92}, the case was made that 
even deductive operations may be too wild, too nonmonotonic, 
and additional monotonicity requirements, termed there rationality
conditions were studied.
We shall study here the strongest of them in its infinitary form.
We shall define Rational Monotonicity and rational operations and
show that, under mild assumptions on \cL, the canonical extension of 
a rational finitary operation is rational.
We shall also provide a representation theorem for rational operations.

\subsection{Rational Operations}
\label{subsec:ratdef}
It seems natural to be most interested in those nonmonotonic inference
operations that are as monotonic as possible. 
One reasonable requirement is that any inference that may be drawn from
a set $X$ may also be drawn from a larger set $Y$, at least if this
larger set is logically consistent with the set of inferences that
could be drawn from $X$. This requirement minimizes the amount of conclusions
that have to be defeated when additional information is gathered.
\begin{definition}
\label{def:rat}
An inference operation \cC\ is {\em rational} iff
it is deductive and satisfies, for any \mbox{$X , Y \subseteq \cL$}
\[\begin{array}{ll}
{\rm (Rational\ Monotonicity)} & \cC(X) \subseteq \cC(X , Y) \\
& \mbox{\rm if } Y \mbox{\rm \ is consistent} \\
& \mbox{\rm with\ } \cC(X)).
\end{array}
\]
A finitary inference operation \cF\ is {\em rational} iff
it is deductive and satisfies, for any \mbox{$A , B \subseteqf \cL$}
\[
{\rm (Finitary\ Rational\ Monotonicity)\ } \ \ \cF(A) \subseteq \cF(A , B) 
{\rm \ if\ } B {\rm \ is\ consistent\ with\ } \cF(A)).
\]
\end{definition}
Notice that we require rational operations to be deductive.
This decision of ours is explained by the fact we do not know of interesting
results concerning cumulative operations that satisfy Rational Monotonicity
but are not deductive.
Notice also that Rational Monotonicity functions as a partial {\em other half}
of the properties defining disjunction and implication.
If \cL\ has disjunction, and if \cC\ satisfies Rational Monotonicity,
then \mbox{$\cC(X , a \vee b) \subseteq$} \mbox{$\cC(X , a) \cap \cC(X , b)$},
at least if both $a$ and $b$ are consistent with \mbox{$\cC(X , a \vee b)$}.
If \cL\ has implication, and if \cC\ satisfies Rational Monotonicity,
then if \mbox{$a \ra b \in \cC(X)$}, then \mbox{$b \in \cC(X , a)$},
at least if $a$ is consistent with \mbox{$\cC(X)$}.
An example of~\cite{Mak:90} shows that some finite Poole systems without 
constraints define inference operations that are not (even finitarily)
rational.
In the framework of classical propositional calculus, rational finitary
operations are exactly the rational relations of~\cite{KLMAI:89} 
and~\cite{LMAI:92}. The reader may find there arguments for finitary rationality.
The following provides a handy characterization of rational operations.
It is closely related to the different equivalent finitary rules of 
restricted transitivity shown to be equivalent to Finitary Rational 
Monotonicity in~\cite{FLMo:91}.
The following theorem provides a characterization of rational operations.
\begin{theorem}
\label{the:ratchar}
An operation \cC\ is rational iff it is supraclassical, left-absorbing
and satisfies \mbox{$\cC(X) = \Cn(X , \cC(Y))$},
for any \mbox{$X , Y \subseteq \cL$} such that
\mbox{$Y \subseteq \cC(X)$} and $X$ is consistent with $\cC(Y)$.
\end{theorem}
\proof
For the {\em only if} part, suppose \cC\ is rational.
It is obviously supraclassical and left-absorbing.
Suppose \mbox{$Y \subseteq \cC(X)$} and 
$X$ is consistent with $\cC(Y)$.
Since \cC\ is deductive and \mbox{$Y \subseteq \cC(X)$},
by Theorem~\ref{the:dedchar}, 
\mbox{$\cC(X) \subseteq \Cn(X , \cC(Y))$}.
By Rational Monotonicity, since $X$ is consistent with $\cC(Y)$, we obtain
\mbox{$\cC(Y) \subseteq \cC(Y , X)$}.
But, since \mbox{$Y \subseteq \cC(X)$}, by Cumulativity,
we have \mbox{$\cC(X) = \cC(X , Y)$} and therefore also
\mbox{$\cC(Y) \subseteq \cC(X)$} and
and \mbox{$\Cn(X , \cC(Y)) \subseteq \cC(X)$}.

For the {\em if} part, suppose \cC\ is supraclassical, left-absorbing and 
that, if
\mbox{$Y \subseteq \cC(X)$} and $X$ is consistent with $\cC(Y)$,
then \mbox{$\cC(X) =$} \nolinebreak[3] \mbox{$\Cn(X , \cC(Y))$}.
We must show that \cC\ is cumulative and satisfies Deductivity and Rational
Monotonicity.
Suppose \mbox{$Y \subseteq \cC(X)$}. 
If $\cC(X)$ is consistent, then, \mbox{$X \cup Y$}, that is a subset of 
$\cC(X)$ is consistent with 
$\cC(X)$ and,
since \mbox{$ X \subseteq \cC(X , Y)$}, we have
\mbox{$\cC(X , Y) =$} \nolinebreak[3] \mbox{$\Cn(X , Y , \cC(X)) =$} 
\nolinebreak[3] \mbox{$\cC(X)$}.
If $\cC(X , Y)$ is consistent, then it is consistent with $X$ and
since \mbox{$X \cup Y$} \nolinebreak[3] \mbox{$\subseteq \cC(X)$} we have
\mbox{$\cC(X) =$} \nolinebreak[3] \mbox{$\Cn(X , \cC(X , Y)) =$} 
\nolinebreak[3] \mbox{$\cC(X , Y)$}.
We are left with the case both $\cC(X)$ and $\cC(X , Y)$ are inconsistent.
But in this case they are both equal to \cL.
We have shown cumulativity.
Let us show Deductivity.
If $X \cup Y$ is consistent with $\cC(Y)$, since \mbox{$Y \subseteq X \cap Y$},
we have \mbox{$\cC(X , Y) =$} \nolinebreak[3] \mbox{$\Cn(X , Y , \cC(Y)) =$}
\nolinebreak[3] \mbox{$\Cn(X , \cC(Y))$}.
If $X \cup Y$ is not consistent with $\cC(Y)$, then 
\mbox{$\cL =$} \nolinebreak[3] \mbox{$\Cn(X , Y , \cC(Y)) =$}
\nolinebreak[3] \mbox{$\Cn(X , \cC(Y))$}
and obviously \mbox{$\cC(X , Y) \subseteq$} \nolinebreak[3] 
\mbox{$\Cn(X , \cC(Y))$}.
We have shown Deductivity.
Let us show Rational Monotonicity.
Suppose that $Y$ is consistent with $\cC(X)$.
Then \mbox{$X \cup Y$} is consistent with $\cC(X)$ and,
since \mbox{$X \subseteq \cC(X , Y)$}, 
\mbox{$\cC(X , Y) =$} \nolinebreak[3] \mbox{$\Cn(X , Y , \cC(X))$}.
Therefore \mbox{$\cC(X) \subseteq$} \nolinebreak[3] \mbox{$\cC(X , Y)$}.
\QED

\subsection{The modular ordering induced by a rational operation}
\label{subsec:modorder}
The relation $\leC$ is not particularly useful in the study of rational
operations. Given an inference operation \cC\, we define a new relation on the
its theories.
\begin{definition}
\label{def:ord2}
Let \cC\ be an inference operation.
Let $X , Y$ be theories.
We shall say that \mbox{$X \lC Y$} iff $X$ is \cC-consistent and
$Y$ is inconsistent with \mbox{$\cC(X \cap Y)$}.
\end{definition}
The relation \mbox{$X \lC Y$} expresses that $X$ is strictly more
expected, or natural, than $Y$. 
If the language \cL\ has disjunction, \mbox{$X \lC Y$} means that $Y$ is
inconsistent with what is expected on the premise that either $X$ or $Y$ holds.
For finitary inference operations, we shall use the following definition,
that assumes the language \cL\ has disjunction.
\begin{definition}
\label{def:ford2}
Assume \cL\ has disjunction.
Let \cF\ be a finitary inference operation.
Let \mbox{$A , B \subseteqf \cL$}.
We shall say that \mbox{$A \lF B$} iff $A$ is \cF-consistent and
$B$ is inconsistent with \mbox{$\cF(A \vee B)$}.
\end{definition}
It is clear that, in definition~\ref{def:ford2}, $A$ could have been replaced
by $\Cn(A)$ and $B$ by $\Cn(B)$; the relation $\lF$ is really a
relation between finitely generated theories.
The following lemma expresses some basic properties of the relation
$\lC$ for a arbitrary cumulative operation \cC.
\begin{lemma}
\label{le:hi2}
Let \cC\ be a cumulative operation.
Let \mbox{$X , Y$} be theories.
\begin{enumerate}
\item \label{irref} the relation \lC\ is irreflexive,
\item \label{asym} if the language \cL\ is admissible,
the relation \lC\ is asymmetric,
\item \label{lCleC} if the language \cL\ is admissible, then 
\mbox{$X \lC Y \Ra X \leC Y$}, 
\item \label{lC} if the language \cL\ is admissible,  
\mbox{$X \lC Y$} iff $X$ is \cC-consistent, \mbox{$X \leC Y$} and 
$Y$ is inconsistent with $\cC(X)$.
\end{enumerate}
\end{lemma}
\proof
Item~\ref{irref} is proved by Inclusion and right-absorption.
For item~\ref{asym}, suppose \mbox{$X \lC Y$} and \mbox{$Y \lC X$}.
We shall derive a contradiction. Let \mbox{$Z \eqdef X \cap Y$}.
We know that both $X$ and $Y$ are \cC-consistent, and both $X$ and $Y$ are 
inconsistent with $\cC(Z)$.
Therefore \mbox{$\cL = \Cn(X , \cC(Z)) \cap \Cn(Y , \cC(Z))$}.
But the language \cL\ is admissible and therefore,
\mbox{$\cL = \Cn(Z , \cC(Z)) = \cC(Z)$}, and $Z$ is \cC-inconsistent.
But, by part~\ref{refl} of Lemma~\ref{le:hi1}, 
\mbox{$Z \leC X$} and by part~\ref{incon} of the same lemma
$X$ is \cC-inconsistent. A contradiction.
For item~\ref{lCleC}, suppose \mbox{$X \lC Y$}. Since $Y$ is inconsistent with
\mbox{$Z \eqdef \cC(X \cap Y)$},
\mbox{$X \subseteq \cL = \Cn(Y , Z)$}.
But \mbox{$X \cap Y \subseteq Z$} and, by Lemma~\ref{le:nadm},
\mbox{$X \subseteq Z$}.
Item~\ref{lC} follows from the previous item and item~\ref{eq} of 
Lemma~\ref{le:hi1}.
\QED
The finitary version of Lemma~\ref{le:hi2} is the following. 
The proof is similar.
\begin{lemma}
\label{le:fhi2}
Assume \cL\ has disjunction.
Let \cF\ be a finitary cumulative operation.
Let \mbox{$A , B$} be finitely generated theories.
\begin{enumerate}
\item \label{firref} the relation \lF\ is irreflexive,
\item \label{fasym} the relation \lF\ is asymmetric,
\item \label{flCleC} \mbox{$A \lF B \Ra A \leF B$}, 
\item \label{flC} \mbox{$A \lF B$} iff $A$ is \cF-consistent, 
\mbox{$A \leF B$} and 
$B$ is inconsistent with $\cF(A)$.
\end{enumerate}
\end{lemma}

One may show that, if \cL\ is admissible and \cC\ is deductive,
the relation \lC\ is a strict partial order, but we shall show directly
a stronger result, i.e. that this relation is a {\em modular} (to be defined) 
partial order if \cC\ is rational.
This result is crucial in the proof of the representation
result of Section~\ref{subsec:ratmod}.
\begin{lemma}
\label{le:modular}
If $\prec$ is a partial order on a set $V$, the following conditions
are equivalent. A partial order satisfying them is called {\em modular}
(this terminology is proposed in~\cite{Gin:86} as an extension of the
notion of modular lattice of~\cite{Gra:71}).
\begin{enumerate}
\item for any \mbox{$x , y , z \in V$} such that \mbox{$x \not\prec y$},
\mbox{$y \not\prec x$} and \mbox{$z \prec x$}, then \mbox{$z \prec y$},
\item  \label{used} for any \mbox{$x , y , z \in V$} such that 
\mbox{$x \prec z$}, either
\mbox{$x \prec y$} or \mbox{$y \prec z$},
\item \label{itnegtrans} the relation $\not\prec$ is transitive,
\item \label{it30} there is a totally ordered set $\Omega$ 
(the strict order on $\Omega$ will be denoted by $<$) 
and a function \mbox{$r : V \mapsto \Omega$} (the ranking function) 
such that 
\mbox{$s \prec t$} iff \mbox{$r(s) < r(t)$}.
\end{enumerate}
\end{lemma}
\begin{lemma}
\label{le:umod}
Any asymmetric binary relation that satisfies property~\ref{used} of 
Lemma~\ref{le:modular} is transitive and therefore a modular strict partial 
order.
\end{lemma}
\proof
Suppose \mbox{$x \prec y \prec z$}.
If we had \mbox{$x \not \prec z$}, since we have \mbox{$x \prec y$},
we would have \mbox{$z \prec y$}, contradicting asymmetry.
\QED
The relation $\lC$ is the principal tool in studying rational operations.
The next two lemmas state important properties of $\lC$ for arbitrary
deductive operations.
\begin{lemma}
\label{le:strictinter}
Let \cC\ be a deductive operation. For theories $X , Y , Z$,
if \mbox{$X \lC Y$}, then \mbox{$X \cap Z \lC Y$}.
\end{lemma}
\proof
Suppose \mbox{$X \lC Y$}. We know that $X$ is \cC-consistent and that
$Y$ is inconsistent with \mbox{$\cC(X \cap Y)$}.
By Lemma~\ref{le:hi1}, part~\ref{refl}, 
\mbox{$X \cap Z \leC X$} and by part~\ref{incon},
\mbox{$X \cap Z$} is \cC-consistent.
We must show that $Y$ is inconsistent with \mbox{$\cC(X \cap Z)$}.
By Deductivity, we have 
\mbox{$\cC(X \cap Y) \subseteq \Cn(X \cap Y , \cC(X \cap Y \cap Z)$}.
But $Y$ is inconsistent with \mbox{$\cC(X \cap Y)$}.
It is therefore inconsistent with \mbox{$\Cn(X \cap Y , \cC(X \cap Y \cap Z)$}.
But \mbox{$\Cn(Y , \Cn(X \cap Y , \cC(X \cap Y \cap Z) =$}
\mbox{$\Cn(Y , \cC(X \cap Y \cap Z)$}.
We conclude that $Y$ is inconsistent with \mbox{$\cC(X \cap Y \cap Z)$}.
\QED
\begin{lemma}
\label{le:<incon}
Let \cC\ be a deductive operation, and $X , Y$ be theories.
If $X$ is \cC-consistent and $Y$ is \cC-inconsistent,
then \mbox{$X \lC Y$}.
\end{lemma}
\proof
Since $X$ is \cC-consistent, we only have to show that $Y$ is
inconsistent with \mbox{$\cC(X \cap Y$}.
But by Deductivity,
\mbox{$\cC(Y) \subseteq \Cn(Y , \cC(X \cap Y)$}.
Since \mbox{$\cC(Y) = \cL$}, we are done.
\QED
The next lemma summarizes the properties of $\lC$ for rational relations.
\begin{lemma}
\label{le:moduC}
Assume \cL\ is admissible.
Let \cC\ be a rational operation.
The relation \lC\ is a modular strict partial order.
\end{lemma}
\proof
By Lemma~\ref{le:hi2}, part~\ref{asym}, \lC\ is asymmetric
(this is where we need the admissibility assumption).
By Lemma~\ref{le:umod}, it enough to show that, 
if \mbox{$X \lC Z$} and \mbox{$X \notlC Y$}, then \mbox{$Y \lC Z$}.
We shall suppose that $X$, $Y$ and $Z$ are theories.
Suppose \mbox{$X \lC Z$} and \mbox{$X \notlC Y$}.
We know that $X$ is \cC-consistent, $Z$ is inconsistent with
\mbox{$\cC(X \cap Z)$} and $Y$ is consistent with
\mbox{$\cC(X \cap Y)$}.
Since $X$ is \cC-consistent and \mbox{$X \notlC Y$},
Lemma~\ref{le:<incon} proves that $Y$ is \cC-consistent.
If $Z$ is \cC-inconsistent, we conclude by Lemma~\ref{le:<incon}.
Assume, then, that $Z$ is \cC-consistent.
We shall now prove that $Z$ is inconsistent
with \mbox{$\cC(Y \cap Z)$}.
We know that \mbox{$\cL = \Cn(Z , \cC(X \cap Z))$}.
By Deductivity, \mbox{$\cC(X \cap Z) \subseteq$} 
\mbox{$\Cn(X \cap Z , \cC(X \cap Y \cap Z))$}.
Therefore we have
\[\begin{array}{lll}
\cL & = & \Cn(Z , \Cn(X \cap Z , \cC(X \cap Y \cap Z))) \\
& = & \Cn(Z , \cC(X \cap Y \cap Z)),
\end{array}
\]
i.e., $Z$ is inconsistent with
\mbox{$\cC(X \cap Y \cap Z)$}.
By Lemma~\ref{le:strictinter}, from \mbox{$X \lC Z$}  we deduce
\mbox{$X \cap Y \lC Z$}. By Lemma~\ref{le:hi2}, part~\ref{asym},
we see that \mbox{$Z \notlC X \cap Y$}.
Since $Z$ is \cC-consistent,
\mbox{$X \cap Y$} is consistent with \mbox{$\cC(X \cap Y \cap Z$}.
By Rational Monotonicity (this the first time we use it in this proof),
we have \mbox{$\cC(X \cap Y \cap Z) \subseteq$} \nolinebreak[3]
\mbox{$\cC(X \cap Y)$}.
But $Y$ is consistent with \mbox{$\cC(X \cap Y)$}, and we conclude
that $Y$ is consistent with \mbox{$\cC(X \cap Y \cap Z)$}.
Therefore \mbox{$Y \cap Z$} is consistent with \mbox{$\cC(X \cap Y \cap Z)$}.
By Rational Monotonicity, now, we have 
\mbox{$\cC(X \cap Y \cap Z) \subseteq$} \nolinebreak[3] \mbox{$\cC(Y \cap Z)$}.
But $Z$ is inconsistent with \mbox{$\cC(X \cap Y \cap Z)$}, it is therefore
inconsistent with \mbox{$\cC(Y \cap Z)$}.
\QED

\subsection{Canonical Extensions of Rational Operations}
\label{subsec:ratcan}
We shall now show that canonical extensions preserve Rational Monotonicity.
The first proof of this result, in the framework of propositional calculus,
was due to David Makinson.
First, we need a technical lemma, concerning arbitrary cumulative operations.
It shows that cumulative operations enjoy a certain measure
of Monotonicity. It may be compared with Theorem~\ref{the:distweak}.
The infinitary version of this lemma holds and is proved similarly.
\begin{lemma}
\label{le:cummon}
Assume the language \cL\ is admissible.
Let \cF\ be a cumulative finitary operation.
For any finitely generated theories \mbox{$A , B , C \subseteq \cL$}, if $C$ 
is consistent with
\mbox{$\cF(A , B)$}, then it is consistent with
\mbox{$\cF(A , B \cap C)$}.
\end{lemma}
\proof
We shall reason by contradiction.
Suppose $C$ is inconsistent with \mbox{$T \eqdef \cF(A , B \cap C)$}.
We have \mbox{$B \subseteq$} \nolinebreak[3] \mbox{$\cL =$} 
\nolinebreak[3] \mbox{$\Cn(T , C)$}.
Since \cL\ is admissible and 
\mbox{$B \cap C \subseteq$} \nolinebreak[3] $T$,
using Lemma~\ref{le:nadm},
we may conclude that
\mbox{$B \subseteq T$}.
By Cumulativity, we have \mbox{$T =$} \nolinebreak[3]
\mbox{$\cF(A , B \cap C , B) =$} \nolinebreak[3]
\mbox{$\cF(A , B)$}.
But $C$ is consistent with \mbox{$\cF(A , B)$}.
A contradiction.
\QED
\begin{theorem}
\label{the:ratext}
Assume the language \cL\ has implication, disjunction
and intuistionistic negation.
Let \cF\ be a rational finitary operation.
The canonical extension of \cF\ is rational and equal to \CF.
\end{theorem}
\proof
In view of Theorem~\ref{the:CFcanext}, we only have to check that
the operation \CF\ satisfies Rational Monotonicity.
Suppose, indeed that $Y$ is consistent with $\CF(X)$.
We shall show that \mbox{$\CF(X) \subseteq \CF(X , Y)$}.
Let $a$ be an arbitrary element of $\CF(X)$.
There is \mbox{$A \subseteqf X$} such that \mbox{$a \in \cF(A , B)$}
for every \mbox{$B \subseteqf \Cn(X)$}.
But \mbox{$A \subseteqf X \cup Y$} and we shall show that
\mbox{$a \in \cF(A , C)$} for every \mbox{$C \subseteqf \Cn(X , Y)$}.
Take an arbitrary such $C$. We notice that $C$ is consistent $\CF(X)$.
We claim, first, that there is \mbox{$D \subseteqf \Cn(X)$}
such that $C$ is consistent with \mbox{$\cF(A , D)$}.
Indeed, if there was no such $D$, for any \mbox{$D \subseteqf \Cn(X)$},
$C$ would be inconsistent with \mbox{$\cF(A , D)$} and, therefore,
\mbox{$\neg \chi_{C} \in \cF(A , D)$}.
By the definition of \CF\ this would imply that
\mbox{$\neg \chi_{C} \in \CF(X)$}, which contradicts the fact
that $C$ is consistent with $\CF(X)$.

Consider, then, such a $D$.
Notice that, by Lemma~\ref{le:basic} (part~\ref{bdis} or part~\ref{bconimp2}), 
\cL\ is admissible and by Lemma~\ref{le:cummon}, 
since $C$ is consistent with \mbox{$\cF(A , D)$},
$C$ is consistent with \mbox{$\cF(A , C \vee D)$}.
By Rational Monotonicity, therefore, 
\mbox{$\cF(A , C \vee D) \subseteq \cF(A , C)$}.
But, since, \mbox{$D \subseteq \Cn(X)$}, we have 
\mbox{$C \vee D \subseteq \Cn(X)$}
and \mbox{$a \in \cF(A , C \vee D)$}.
\QED

\subsection{Modular Models and Rational Operations}
\label{subsec:ratmod}
We shall now describe a family of full models that corresponds exactly
to rational operations.
The representation result is the infinitary version of the corresponding 
result of~\cite{LMAI:92}. 
\begin{definition}
\label{def:rankmod}
A full model \mbox{$W \eqdef \langle S , l , \prec \rangle$} for which the 
strict partial order $\prec$ is modular is said to be a modular model.
\end{definition}
Notice that our modular models are similar to the ranked models of~\cite{LMAI:92}
but differ from them in two respects: the smoothness property we require from
modular models is stronger and we require them to be full.
The next soundness result requires no assumption on \cL.
\begin{theorem}
\label{the:soundrat}
If $W$ is a modular model, the inference operation \CW\ it defines
is rational.
\end{theorem}
\proof
By Theorem~\ref{the:soundfull} it is enough to show that \CW\ satisfies 
Rational Monotonicity.
Suppose $W$ is a modular model.
We shall use the notations of 
Definition~\ref{def:rankmod}. Suppose also that $Y$ is consistent with
$\CW(X)$.
Since $W$ is full, we conclude that there is a minimal element 
\mbox{$s \in S$} of $\widehat{X}$ that satisfies $Y$.
Let $t$ be any minimal element of $\widehat{X \cup Y}$.
Clearly $t$ is an element of $\widehat{X}$. We claim it is
a minimal element of $\widehat{X}$. Indeed, if it is not the case,
by the smoothness condition, there is a state \mbox{$v \prec t$}
that is minimal in $\widehat{X}$. But, since $s$ is minimal in $\widehat{X}$,
\mbox{$v \not\prec s$} and, by modularity, \mbox{$s \prec t$}.
But $s$ is in $\widehat{X \cup Y}$ and $t$ is minimal in this set.
A contradiction.
We have shown that any minimal element of $\widehat{X \cup Y}$ is a minimal
element of $\widehat{X}$. Therefore \mbox{$\CW(X) \subseteq \CW(X , Y)$}.
\QED
We want, now, to prove the converse of Theorem~\ref{the:soundrat}.
The method of proof we propose is an improvement on that of~\cite{LMAI:92}.
The representation theorem that results is also a strengthening of the
corresponding result there (see our comments at the end of 
Section~\ref{subsec:cummod}).
We shall assume that \cL\ is admissible.
Suppose a rational operation \cC\ is given.
Notice that, by Theorem~\ref{the:deddist}, \cC\ is distributive.
We shall define the structure $W$ to be the triple 
\mbox{$\langle S , l , \prec \rangle$}, where 
\begin{itemize}
\item $S$ is the set of all pairs
\mbox{$(m , X)$} where $m$ is a normal world for $X$, 
i.e. \mbox{$m \models \cC(X)$}, and $m$ does not satisfy all the formulas
of \cL,
\item $l$ is the projection on the
first coordinate, i.e. \mbox{$l(m , X) = m$} and 
\item the relation $\prec$
is defined by: \mbox{$(m , X) \prec (n , Y)$} iff \mbox{$X \lC Y$}.
\end{itemize}
Notice that, if \mbox{$(m , X)$} is a state, then $X$ is \cC-consistent,
since $m$ is normal for $X$ but does not satisfy \cL.
Our first task is to show that $W$ is a modular model: we must show it 
satisfies the smoothness condition, the fullness condition and that $\prec$
is a modular strict partial order.
Then, we shall show that it defines the operation \cC.
\begin{lemma}
\label{le:modo}
The binary relation $\prec$ on $S$ is a modular strict partial order.
\end{lemma}
\proof
Obvious from Lemma~\ref{le:moduC}.
\QED
\begin{lemma}
\label{le:minichar}
\begin{enumerate}
\item If $Y$ is \cC-consistent,
a state \mbox{$(m , X)$} of $S$ is minimal in $\widehat{Y}$ iff
\mbox{$m \models Y$} and \mbox{$Y \notlC X$}.
\item If $Y$ is \cC-inconsistent, the set $\widehat{Y}$ is empty.
\end{enumerate}
\end{lemma}
\proof
Let, first, $Y$ be \cC-consistent.
Notice that, by the completeness of the semantics, there is a world $n$ 
that is normal for $Y$ and does not satisfy all formulas of \cL. 
The pair \mbox{$(n , Y)$} is a state of $S$.
Suppose first that \mbox{$(m , X)$} is minimal in $\widehat{Y}$.
By the definition of $l$, $m$ satisfies $Y$.
If we had \mbox{$Y \lC X$}, we would have \mbox{$(n , Y) \prec (m , X)$}
for a state \mbox{$(n , Y)$} that satisfies $Y$, contradicting the minimality 
of \mbox{$(m , X)$}. Therefore \mbox{$Y \notlC X$}.

Suppose now that \mbox{$(m , X) \in \widehat{Y}$} and that 
\mbox{$Y \notlC X$}.
If there was a state \mbox{$(n , Z) \prec (m , X)$} such that
\mbox{$n \models Y$}, we would have \mbox{$Z \lC X$}, and $n$ would satisfy 
both $\cC(Z)$ and $Y$ and, since it does not satisfy all formulas of \cL,
$Y$ would be consistent with $\cC(Z)$ and therefore we would have
\mbox{$Z \notlC Y$}.
But, we would have \mbox{$Z \notlC Y \notlC X$} and, since
\lC\ is modular, by Lemma~\ref{le:moduC}, 
we would have, by item~\ref{itnegtrans} of Lemma~\ref{le:modular},
\mbox{$Z \notlC X$}. A contradiction.

Let, now, $Y$ be \cC-inconsistent. Let $X$ be any set of formulas.
Since \mbox{$\cL =$} \nolinebreak[3] \mbox{$\cC(Y)$}, 
we have, by Cautious Monotonicity,
\mbox{$\cL =$} \nolinebreak[3] \mbox{$\cC(Y , X)$} and, by Deductivity, 
we obtain
\mbox{$\cL =$} \nolinebreak[3] \mbox{$\Cn(Y , \cC(X))$}.
Any normal world for $X$ that satisfies $Y$ satisfies all the formulas
of \cL. There is therefore no state in $\widehat{Y}$.
\QED
\begin{lemma}
\label{le:smoothness}
The structure $W$ satisfies the smoothness condition.
\end{lemma}
\proof
Let \mbox{$(m , X)$} be in $\widehat{Y}$.
Either \mbox{$Y \notlC X$} and \mbox{$(m , X)$} is minimal in 
$\widehat{Y}$,
by Lemma~\ref{le:minichar},
or \mbox{$Y \lC X$}.
In such a case, $Y$ is \cC-consistent and there is a normal world $n$ for $Y$
and we have \mbox{$(n , Y) \prec (m , X)$}.
But \mbox{$(n , Y) $} is, by Lemma~\ref{le:minichar}, a minimal state in 
$\widehat{Y}$.
\QED
\begin{lemma}
\label{le:ratCWC}
The operations \cC\ and \CW\ coincide.
\end{lemma}
\proof
If \mbox{$a \in \CW(Y)$}, $a$ is satisfied in every world $m$ that is normal
for $Y$, since for such a world \mbox{$(m , Y)$} is a minimal state of
$\widehat{Y}$. Therefore, we have \mbox{$a \in \Cn(\cC(Y)) =$} 
\nolinebreak[3] \mbox{$\cC(Y)$}.
Suppose, now, that \mbox{$a \in \cC(Y)$}.
Let \mbox{$(n , X)$} be minimal in $\widehat{Y}$.
By Lemma~\ref{le:minichar} \mbox{$Y \notlC X$}. 
If $Y$ is \cC-inconsistent, then $\widehat{Y}$ is empty by 
Lemma~\ref{le:minichar}
and \mbox{$\CW(Y) = \cL$}.
If $Y$ is \cC-consistent, then, $X$ is consistent with 
\mbox{$\cC(\Cn(X) \cap \Cn(Y))$}.
Therefore, by Rational Monotonicity,
\mbox{$\cC(\Cn(X) \cap \Cn(Y)) \subseteq$} \nolinebreak[3] \mbox{$\cC(X)$}
But, by Deductivity,
\mbox{$\cC(Y) \subseteq$} \nolinebreak[3] 
\mbox{$\Cn(Y , \cC(\Cn(X) \cap \Cn(Y)))$}.
We conclude that \mbox{$\cC(Y) \subseteq$} \nolinebreak[3] 
\mbox{$\Cn(Y , \cC(X))$}.
Since $n$ satisfies both
$\cC(X)$ and $Y$ it satisfies $a$.
We conclude that \mbox{$a \in \CW(Y)$}.
\QED
\begin{lemma}
\label{le:full}
The structure $W$ satisfies the fullness property.
\end{lemma}
\proof
If $m$ satisfies all the formulas of $\CW(X)$, by Lemma~\ref{le:ratCWC},
it is normal for $X$. If $m$ does not satisfy all the formulas of \cL,
the pair \mbox{$(m , X)$} is a state in $W$ and it is minimal in
$\widehat{X}$.
\QED
We may now conclude.
\begin{theorem}
\label{the:ratrep}
Assume \cL\ is admissible.
An inference operation \cC\ is rational iff it is defined by some modular
model.
\end{theorem}
\proof
By Theorem~\ref{the:soundrat} and the construction just above.
\QED

\section{Conclusions and Further Work}
\label{sec:conc}
We have presented a number of families of nonmonotonic inference operations.
We have studied the question of extending finitary operations to infinitary
ones, and obtained representation results.
Most of the insightful ideas about infinitary operations have come from
the study of finitary operations. 
The first benefit gained from such a study of infinitary operations is 
a streamlining of the results and the proofs.
But, now, time is perhaps ripe for the study of infinitary operations 
to suggest new ideas concerning finitary operations. 
One such idea seems particularly attractive.
Since we have seen two incomparable families of finitary operations \cF\
that have the property that \CF\ is equal to its transform $\CF'$,
namely finitary distributive operations (if \cL\ has disjunction),
on the one hand, and finitary operations defined by finite Poole systems
(if \cL\ has contradiction) on the other hand, the family of finitary
operations that possess this property is probably an interesting object of 
study.

A novel aspect of this work, compared with all previous works on general
patterns of nonmonotonic reasoning is its abstract handling of the
operation of logical consequence \Cn.
We are not sure that the assumptions made on \cL\ are necessary.
It would be interesting to study exotic consequence operations,
that have very few connectives,
and see what holds true there.
\section*{Acknowledgements}
\label{sec:ack}
David Makinson was with us all along during the elaboration of this
work. We tried to credit him in the text with the results that are
definitely his, but his influence on this paper goes further.
We also want to thank Karl Schlechta and Zeev Geyzel for their remarks
that lead to definite improvements.

\appendix
\section{Properties of the language \cL}
\label{app:cL}
Here, we shall define properties of the language \cL\ (more precisely,
the language \cL\ and the consequence operation \Cn), that are needed
at different stages in our work. We shall also mention some technical
lemmas. The Appendix is self-contained and does not rely on any
result appearing in the main part of the paper.
We shall first describe a property of \Cn\ that we have found to play a 
fundamental role in the study of certain families of inference operations.
This property is not enjoyed by all consequence operations. It follows
from the existence of different sets of connectives, but may be true even for 
languages that do not possess those connectives.
\begin{definition}
\label{def:adm}
The language \cL\ is {\em admissible} iff,
for any theories $X$, $Y$, $Z$,
\begin{equation} \label{eq:ad}
\Cn(X , Y) \cap \Cn(X , Z) = \Cn(X , Y \cap Z).
\end{equation}
\end{definition}
Notice that, since it is always the case that
\begin{equation}
\label{eq:half} 
\Cn(X , Y \cap Z) \subseteq \Cn(X , Y) \cap \Cn(X , Z),
\end{equation}
Equation~(\ref{eq:ad}) could have been replaced by the inclusion from
left to right. Admissibility is, for \Cn, the property of 
Distributivity that is defined in a more general setting 
in Section~\ref{subsec:basic}, 
Equation~(\ref{eq:dis}).
We shall see, in Lemma~\ref{le:basic}, that the existence
of certain connectives ensures admissibility,
but one should notice that, if \Cn\ is taken to be the identity,
though no connectives are present, the language is admissible.
An example of a language that is not admissible is obtained by
considering propositional variables and negated propositional variables
and defining $\Cn(X)$ as \cL\ if $X$ contains some variable and its
negation and $X$ otherwise.

We shall now define properties of \cL\ and \Cn\ corresponding to the existence
of well-behaved connectives.
\begin{definition}
\label{def:connectives}
The language \cL\ is said 
\begin{enumerate}
\item to have {\em contradiction} iff there is a formula {\bf false}
such that \mbox{$\Cn({\bf false}) = \cL$},
\item to have {\em conjunction} iff, for any formulas \mbox{$a , b \in \cL$},
there is a formula \mbox{$a \wedge b$} such that, for any 
\mbox{$X \subseteq \cL$}, \mbox{$\Cn(X , a , b) = \Cn(X , a \wedge b)$},
\item to have {\em disjunction} iff, for any formulas \mbox{$a , b \in \cL$},
there is a formula \mbox{$a \vee b$} such that, for any 
\mbox{$X \subseteq \cL$}, \mbox{$\Cn(X , a \vee b) = 
\Cn(X , a) \cap \Cn(X , b)$},
\item to have {\em implication} iff, for any formulas \mbox{$a , b \in \cL$},
there is a formula \mbox{$a \ra b$} such that, for any 
\mbox{$X \subseteq \cL$}, \mbox{$b \in \Cn(X , a)$} iff 
\mbox{$a \ra b \in \Cn(X)$},
\item to have {\em classical negation} iff, for any formula \mbox{$a \in \cL$},
there is a formula \mbox{$\neg a$} such that, for any 
\mbox{$X \subseteq \cL$}, \mbox{$a \in \Cn(X)$} iff 
\mbox{$\Cn(X , \neg a) = \cL$},
\item to have {\em intuistionistic negation} iff, for any formula 
\mbox{$a \in \cL$},
there is a formula \mbox{$\neg a$} such that, for any 
\mbox{$X \subseteq \cL$}, \mbox{$\neg a \in \Cn(X)$} iff 
\mbox{$\Cn(X , a) = \cL$}.
\end{enumerate}
\end{definition}
Notice that both classical and intuitionistic propositional calculi
have contradiction, conjunction, disjunction and implication.
Intuistionistic propositional calculus has intuitionistic negation.
Classical propositional calculus has classical (and intuistionistic)
negation.

If \cL\ has disjunction and \mbox{$X , Y \subseteq \cL$} we shall denote
by \mbox{$X \vee Y$}, the set of all formulas \mbox{$x \vee y$} for
\mbox{$x \in X , y \in Y$}.
If \cL\ has conjunction and \mbox{$A \subseteqf \cL$}, we shall denote
the characteristic formula of $A$, i.e., the conjunction of all the formulas 
of $A$ by $\chi_{A}$, i.e., \mbox{$\chi_{A} \eqdef \bigwedge_{a \in A} a$}.
The following long lemma summarizes classical results we shall need further.
The proofs will either be skipped or only sketched.
\begin{lemma}
\label{le:basic}
\ 
\begin{enumerate}
\item \label{bcontr}
If the language \cL\ has contradiction and $X$ is inconsistent, then there is
a finite subset $A$ of $X$ that is inconsistent.
\item \label{bdis}
If the language \cL\ has disjunction,
for any sets $X$, $Y$, $Z$ of formulas, 
\mbox{$\Cn(X , Y \vee Z) = $}
\mbox{$\Cn(X , Y) \cap \Cn(X , Z)$}.
In particular, we have
\mbox{$\Cn(X \vee Y) =$} \nolinebreak[3] \mbox{$\Cn(X) \cap \Cn(Y)$} and
the language \cL\ is admissible.
\item \label{bconimp1}
If the language \cL\ has implication,
for any set $Y$ and any finite set $A$ of formulas,
there is a set \mbox{$A \ra Y$} of formulas such that,
for any set $X$ of formulas,
\mbox{$A \ra Y \subseteq \Cn(X)$} iff \mbox{$Y \subseteq \Cn(X , A)$}.
If $Y$ is finite, so is \mbox{$A \ra Y$}.
\item \label{bconimp2}
If the language \cL\ has implication, then it is admissible.
\item \label{bconimp3}
If the language \cL\ has implication,
for any finite set $A$ of formulas and any family 
\mbox{$Y_{i} , i \in I$} of sets of formulas,
\mbox{$\Cn(A , \bigcap_{i \in I} \Cn(Y_{i})) = $} \nolinebreak[3]
\mbox{$\bigcap_{i \in I} \Cn(A , Y_{i})$}.
\item \label{bclassneg}
If the language \cL\ has classical negation,
it has intuistionistic negation and, for any formula $a$, 
\mbox{$a \in \Cn(\neg \neg a)$}.
\item \label{bnewclassneg}
If the language \cL\ has conjunction and classical negation,
for any finite set $A$ of formulas, 
\mbox{$\Cn(A) \cap \Cn( \neg \chi_{A}) =$} \mbox{$\Cn(\emptyset)$}.
\item \label{bintneg}
If the language \cL\ has intuistionistic negation,
for any formula $a$, \mbox{$\Cn(a , \neg a) =$} \nolinebreak[3] $\cL$
and \mbox{$\neg \neg a \in \Cn(a)$}. 
\end{enumerate}
\end{lemma}
\proof
Property~\ref{bcontr} is easy to prove, by the compactness of \Cn.
Notice that we do not assume, in property~\ref{bdis} that \cL\ has conjunction.
First, we shall prove the equality when the sets $Y$ and $Z$ are finite.
For that, we proceed by induction on the size of the set \mbox{$Y \vee Z$}.
If it is empty, then one of the sets $Y$ or $Z$ is empty and the result is
proven.
Suppose \mbox{$Y = Y' \cup \{ y \}$}.
Then, \mbox{$\Cn(X , Y \vee Z) = \Cn(X , Y' \vee Z , \{ y \} \vee Z)$}.
Since \cL\ has disjunction,
\[
\Cn(X , Y' \vee Z , \{ y \} \vee Z) = \Cn(X , Y' \vee Z , y)
\cap \Cn(X , Y' \vee Z , Z).
\]
We conclude by the induction hypothesis.
For the general case, when $Y$ and $Z$ may be infinite, first, 
notice that the inclusion from left to right is easily proven.
Then, use compactness, apply what has just been proved when $Y$ and $Z$
are finite and conclude by Monotonicity.
The special case is obtained for $X$ empty.
It follows that \cL\ satisfies Equation~(\ref{eq:ad}).
For property~\ref{bconimp1}, take the set \mbox{$A \ra Y$} to be the set of 
all formulas of the form
\mbox{$a_{0} \ra a_{1} \ra \ldots \ra a_{n} \ra y$} for \mbox{$y \in Y$},
and \mbox{$\{ a_{0} , a_{1} , \ldots a_{n} \}$} is an enumeration of $A$.
For property~\ref{bconimp2},
one inclusion is obvious, as noticed following Definition~\ref{def:adm}.
Suppose now that \mbox{$a \in \Cn(X , Y) \cap \Cn(X , Z)$}.
By compactness, there is \mbox{$A \subseteqf X$} such that
\mbox{$a \in \Cn(A , Y) \cap \Cn(A , Z)$}.
Since \cL\ has implication, by property~\ref{bconimp1}
we have
\mbox{$A \ra \{ a \} \subseteq \Cn(Y) \cap \Cn(Z)$}.
Therefore \mbox{$a \in \Cn(A , \Cn(Y) \cap \Cn(Z))$}.
One concludes by Monotonicity.
For property~\ref{bconimp3}, as above
the inclusion from left to right is obvious.
Suppose now that \mbox{$a \in \bigcap_{i \in I} \Cn(A , Y_{i})$}.
Then one has, by property~\ref{bconimp1}, 
\mbox{$A \ra \{ a \} \subseteq \bigcap_{i \in I} \Cn(Y_{i})$}
and one concludes easily.
Properties~\ref{bclassneg}, \ref{bnewclassneg} and~\ref{bintneg} are easily 
proven.
\QED

The following is a technical lemma needed in Sections~\ref{subsec:dedrep},
\ref{subsec:modorder} and~\ref{subsec:ratcan}.
\begin{lemma}
\label{le:nadm}
Assume the language \cL\ is admissible. For any theories $X$, $Y$, $Z$,
if \mbox{$X \subseteq \Cn(Z , Y)$} and 
\mbox{$X \cap Y \subseteq Z$}, then \mbox{$X \subseteq Z$}.
\end{lemma}
\proof
Suppose \mbox{$X \subseteq$} \nolinebreak[3] \mbox{$\Cn(Z , Y)$} and 
\mbox{$X \cap Y \subseteq Z$}.
From the second hypothesis we see that
\mbox{$\Cn(Z , X \cap Y) \subseteq Z$}.
Since \cL\ is admissible,
\mbox{$\Cn(Z , X) \cap \Cn(Z , Y) \subseteq$} \nolinebreak[3]
\mbox{$\Cn(Z , X \cap Y)$}.
Therefore
\mbox{$\Cn(Z , X) \cap \Cn(Z , Y) \subseteq Z$}.
But $X$ is clearly a subset of the intersection on the left.
\QED

\bigskip

\noindent Received 30 April 1993.

\begin{thebibliography}{10}

\bibitem{DM:92}
J\"{u}rgen Dix and David Makinson.
\newblock The relationship between KLM and MAK models for nonmonotonic inference operations.
\newblock {\em Journal of Logic, Language and Information}, 1(2):131--140, 1992.

\bibitem{NIL:90}
Michael Freund.
\newblock Supracompact inference operations.
\newblock In J.~Dix, K.~P. Jantke, and P.H. Schmitt, editors, {\em Nonmonotonic
  and Inductive Logic, First International Workshop, Karlsruhe Germany}, pages
  59--73. Springer Verlag, December 1990.
\newblock Lecture Notes in Artificial Intelligence, Vol. 543.

\bibitem{Freund:92}
Michael Freund.
\newblock Supracompact inference operations.
\newblock Submitted, January 1992.

\bibitem{FL:JELIA90}
Michael Freund and Daniel Lehmann.
\newblock Deductive inference operations.
\newblock In J.~van Eijck, editor, {\em European Workshop on Logical
Methods in Artificial Intelligence, Amsterdam the Netherlands}, pages 227--233.
Springer Verlag, September 1990.
\newblock Lecture Notes in Artificial Intelligence, Vol. 478.

\bibitem{FLM:89}
Michael Freund, Daniel Lehmann, and David Makinson.
\newblock Canonical extensions to the infinite case of finitary nonmonotonic
  inference operations.
\newblock In {\em Workshop on Nomonotonic Reasoning}, pages 133--138, Sankt
  Augustin, FRG, December 1989. Arbeitspapiere der GMD no. 443.

\bibitem{FLMo:91}
Michael Freund, Daniel Lehmann, and Paul Morris.
\newblock Rationality, transitivity, and contraposition.
\newblock {\em Artificial Intelligence}, 52(2):191--203, December 1991.
\newblock Research Note.

\bibitem{Gabbay:85}
Dov~M. Gabbay.
\newblock Theoretical foundations for non-monotonic reasoning in expert
  systems.
\newblock In Krzysztof~R. Apt, editor, {\em Proc. of the NATO Advanced Study
  Institute on Logics and Models of Concurrent Systems}, pages 439--457, La
  Colle-sur-Loup, France, October 1985. Springer-Verlag.

\bibitem{Gent:69}
Gerhard Gentzen.
\newblock {\em The Collected Papers of Gerhard Gentzen, edited by M. E. Szabo}.
\newblock North Holland, Amsterdam, 1969.

\bibitem{Gin:86}
Matthew~L. Ginsberg.
\newblock Counterfactuals.
\newblock {\em Artificial Intelligence}, 30:35--79, 1986.

\bibitem{Gra:71}
George Gr\"{a}tzer.
\newblock {\em Lattice Theory}.
\newblock W. H. Freeman, San Francisco, 1971.

\bibitem{KLMAI:89}
Sarit Kraus, Daniel Lehmann, and Menachem Magidor.
\newblock Nonmonotonic reasoning, preferential models and cumulative logics.
\newblock {\em Artificial Intelligence}, 44(1--2):167--207, July 1990.

\bibitem{Leh:89}
Daniel Lehmann.
\newblock What does a conditional knowledge base entail?
\newblock In Ron Brachman and Hector Levesque, editors, {\em Proceedings of the
  First International Conference on Principles of Knowledge Representation and
  Reasoning}, Toronto, Canada, May 1989. Morgan Kaufmann.

\bibitem{Leh:91}
Daniel Lehmann.
\newblock Plausibility logic.
\newblock In Egon Boerger, Gerhard J\"{a}ger, Hans~Kleine Buening, and
  Michael~M. Richter, editors, {\em Proceedings of Computer Science Logic '91},
  volume 626, pages 227--241, Berne, Switzerland, October 1991. Lecture Notes
  in Computer Science, Springer Verlag.

\bibitem{LMTARK:90}
Daniel Lehmann and Menachem Magidor.
\newblock Preferential logics: the predicate calculus case.
\newblock In Rohit Parikh, editor, {\em Proceedings of the Third Conference on
  Theoretical Aspects of Reasoning About Knowledge}, pages 57--72, Monterey,
  California, March 1990. Morgan Kaufmann.

\bibitem{LMAI:92}
Daniel Lehmann and Menachem Magidor.
\newblock What does a conditional knowledge base entail?
\newblock {\em Artificial Intelligence}, 55(1):1--60, May 1992.

\bibitem{Mak:89}
David Makinson.
\newblock General theory of cumulative inference.
\newblock In M.~Reinfrank, J.~de~Kleer, M.~L. Ginsberg, and E.~Sandewall,
  editors, {\em Proceedings of the Second International Workshop on
  Non-Monotonic Reasoning}, pages 1--18, Grassau, Germany, June 1988. Springer
  Verlag.
\newblock Volume 346, Lecture Notes in Artificial Intelligence.

\bibitem{Mak:90}
David Makinson.
\newblock General patterns in nonmonotonic reasoning.
\newblock In D.~M. Gabbay, C.~J. Hogger, and J.~A. Robinson, editors, {\em
  Handbook of Logic in Artificial Intelligence and Logic Programming Vol. 2,
  Nonmonotonic and Uncertain Reasoning}. Oxford University Press, due 1992.
\newblock in preparation.

\bibitem{Poole:88}
David Poole.
\newblock A logical framework for default reasoning.
\newblock {\em Artificial Intelligence}, 36:27--47, 1988.

\bibitem{SchleInf:91}
Karl Schlechta.
\newblock Results on infinite extensions.
\newblock {\em Journal of Applied Non-Classical Logics}, 1(1):65--72, 1991.

\bibitem{Schle:91}
Karl Schlechta.
\newblock Some results on classical preferential models.
\newblock {\em Journal of Logic and Computation}, 2(6):in print, 1992.

\bibitem{Tar:56}
Alfred Tarski.
\newblock {\em Logic, Semantics, Metamathematics. Papers from 1923--1938}.
\newblock Clarendon Press, Oxford, 1956.

\end{thebibliography}
\end{document}